\documentclass{article}
\usepackage{float}
\usepackage{amsmath,amssymb,amsthm}
\usepackage{iclr2027_conference,times}
\usepackage{graphicx} 
\usepackage{adjustbox}

\usepackage{amsmath,amsfonts,bm}

\def\eqref#1{equation~\ref{#1}}

\def\1{\bm{1}}

\DeclareMathAlphabet{\mathsfit}{\encodingdefault}{\sfdefault}{m}{sl}
\SetMathAlphabet{\mathsfit}{bold}{\encodingdefault}{\sfdefault}{bx}{n}

\usepackage{xcolor}

\definecolor{cemgray}{RGB}{245,245,245}
\definecolor{tierorange}{RGB}{255,237,210}
\definecolor{tierblue}{RGB}{225,230,250}
\definecolor{fallbackpink}{RGB}{255,225,235}
\definecolor{elitegreen}{RGB}{225,245,230}
\definecolor{cemgraystrong}{HTML}{606060}
\definecolor{tierorangestrong}{HTML}{eda43c}
\definecolor{tierbluestrong}{HTML}{4d50ad}
\definecolor{fallbackpinkstrong}{HTML}{c80e47}
\definecolor{elitegreenstrong}{HTML}{2c8a5a}

\usepackage{tcolorbox}

\makeatletter
\newcommand{\algshade}[2]{%
    \State
    \colorbox{#1}{%
        \parbox{\dimexpr\linewidth - \ALG@tlm - 2\fboxsep\relax}{#2}%
    }%
}
\makeatother

\usepackage{soul}
\usepackage{caption}
\usepackage{makecell}
\usepackage{booktabs}
\usepackage{subcaption}
\usepackage{tikz}
\usetikzlibrary{shapes.geometric, arrows.meta, positioning, fit, backgrounds, calc}

\usepackage{svg}
\usepackage{import}
\usepackage{hyperref}
\usepackage{url}
\usepackage{doi}
\hypersetup{colorlinks=true, citecolor=gray, linkcolor=gray, urlcolor=gray}
\usepackage{algorithm}
\usepackage{algpseudocode}
\usepackage{footmisc}
\usepackage{placeins}

\usepackage{empheq}
\usepackage{annotate-equations}

\tikzset{annotate equations/arrow/.style={-, shorten <=2pt}}
\usepackage{standalone}
\tikzset{
  every node/.append style={font=\sffamily}
}
\usetikzlibrary{arrows.meta}

\usepackage{amsthm}
\newtheorem{remark}{Remark}
\usepackage{comment}
\makeatletter
\renewcommand{\eqref}[1]{\textup{\tagform@{\ref{#1}}}}
\makeatother

\title{Cross-entropy optimization \\with prioritized constraints}
\author{\ReferenceAuthor{Francisco Roldan Sanchez}{1},
\ReferenceAuthor{Pau de las Heras Molins}{1},\\
\ReferenceAuthor{David Fridovich-Keil}{2}, and
\ReferenceAuthor{Georgios Bakirtzis}{1}}
\ReferenceAffiliations{%
  \ReferenceAffiliation{1}{LTCI, Télécom Paris, Institut Polytechnique de Paris}\\
  \ReferenceAffiliation{2}{The University of Texas at Austin}}
\begin{document}
\maketitle

\begin{abstract}
When constraints conflict, an optimizer must determine which requirements to preserve and which to relax. On the one hand, a priority ordering specifies which requirements take precedence. On the other hand, penalty-based formulations encode their relative importance through numerical weights. Depending on these weights, a solution can improve its weighted score while violating intended priorities. We introduce TierCEM, a variant of the cross-entropy method that incorporates strict constraint priorities directly into elite selection without requiring per-constraint importance weights. TierCEM works by sequentially filtering sampled candidates, from highest- to lowest-priority constraint. If and when a constraint eliminates all remaining candidates, TierCEM returns to the last nonempty set and selects elites with the smallest violations of that blocking constraint, recursively preserving satisfaction of all higher-priority constraints. We evaluate TierCEM on 2D navigation and contact-rich pushing tasks in proprioceptive and learned world-model settings. Experiments show that reversing the constraint ordering changes which constraints are violated under conflict. Prioritizing progress toward the task objective also enables TierCEM to relax lower-priority constraints when they would otherwise prevent further progress.
\end{abstract}

\section{Introduction}

When constraints have different priorities, optimizing a scalarized objective turns the priority order into a weighted trade-off and the optimizer can sacrifice a higher-priority constraint to reduce lower-priority violations. In autonomous driving, for example, better lane keeping must not compensate for a collision when collision avoidance takes precedence~\citep{censi2019nutonomy}. A priority ordering among possibly conflicting constraints makes that distinction explicit. Satisfying a lower-priority requirement must not sacrifice a higher-priority requirement that could otherwise be preserved. %

Incorporating multiple constraints into a single optimization problem is often achieved by combining the task objective and constraint penalties into a scalar score, making constraint priorities dependent on numerical trade-offs between the competing terms~\citep{ma2022conservative, hu2023evolving, kushwaha2025survey}. Under that task objective, an improvement in objective value or lower-priority penalties can compensate for a higher-priority violation. Preventing such compensation requires weights calibrated to both the prescribed ordering and the scales and attainable values of the competing terms, making constraint satisfaction sensitive to that calibration~\citep{imanov2026autonomous}. %

\begin{figure}[!t]
\centering
\begingroup
    \sffamily
    \def\svgwidth{\columnwidth}
    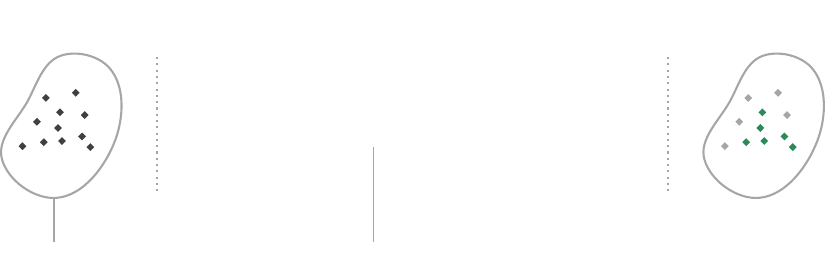
\endgroup
\caption{At each iteration, TierCEM samples candidate trajectories from a Gaussian distribution and filters them from highest- to lowest-priority constraint, producing nested candidate sets $\mathcal{K}_n$. If a tier yields an empty set ($\mathcal{K}_{n^*}=\emptyset$), TierCEM returns to the preceding nonempty set $\mathcal{K}_{n^*-1}$ and selects elites with the smallest violations of the constraint at tier $n^*$, preserving all higher-priority constraints already enforced. Otherwise, elites are selected from the final set according to the smallest violations of the constraint at tier $N$. The elites update the Gaussian sampling distribution for the next iteration; at the final iteration, a plan is selected for execution.}
\label{fig:tiercem}
\end{figure}

Sampling-based optimizers such as the cross-entropy method (CEM)~\citep{botev2013cross} and model-predictive path integral (MPPI)~\citep{williams2017model} provide a natural setting for candidate-wise constraint evaluation. In standard CEM, the top $\ell$ candidates according to a ranking metric are called elites and are used to update the sampling distribution in an iterative fashion. We propose the tiered CEM (TierCEM), which implements ordered constraint relaxation directly in CEM's elite selection logic through a priority-relaxation cascade (cf. Figure~\ref{fig:tiercem}). At each iteration, TierCEM filters sampled candidate solutions from highest- to lowest-priority constraint. If a filter eliminates every remaining candidate, the cascade returns to the preceding nonempty elite set and selects samples with the smallest violations of the blocking constraint. %
If all filters retain candidates, those that maximize the distance to the boundary of the lowest-priority constraint are selected as elites. 

The selected elites then determine the next sampling distribution through the standard CEM update. The constraint ordering thus governs which requirements are preserved and which violation guides relaxation when the current sample batch contains no jointly satisfying candidate, without per-constraint importance weights. 

We evaluate TierCEM in the context of trajectory optimization, where different constraint orderings lead to distinct behavioral tradeoffs.
In a controlled two-dimensional navigation problem with explicit dynamics, geometric conflicts prevent simultaneous satisfaction of all spatial constraints while also reaching the goal. Reversing the constraint ordering changes which spatial constraint is violated. 
We then evaluate contact-rich manipulation in the Push-T environment~\citep{chi2024diffusion}, optimizing action sequences through a frozen latent-space world model and evaluating constraints with learned decoder heads that predict constraint violation.
World model-based experiments reproduce the priority-dependent relaxation observed in the controlled setting: changing the ordering changes which constraint is violated, even when future states and constraint satisfaction are evaluated using imperfect learned predictions.

\section{Related work }

\paragraph{Sampling-based optimization with constraints.}
Constrained CEM can separate constraint satisfaction from task reward when selecting the candidates that update the sampling distribution~\citep{wen2018constrained}. Extensions incorporate differentiable optimization~\citep{mottahedi2022constrained}, while related constrained optimization methods adjust feasibility tolerances through adaptive $\epsilon$-constraint handling~\citep{liu2024dynamic}. Within model-based optimization, \citet{liu2020constrained} use ensemble predictions to evaluate safety, selecting feasible candidates by reward and falling back to the lowest-cost candidates when no feasible sample remains. Their fallback ranks violating candidates by an accumulated constraint cost. TierCEM instead uses an ordering over individual constraints to determine which violation guides elite selection and which higher-priority requirements the selected candidates must preserve. The distinction is therefore not the existence of an infeasibility fallback, but how the prescribed priorities govern that fallback and the resulting distribution update.

\paragraph{Lexicographic and hierarchical constraint handling.}
Lexicographic optimization specifies precedence among objectives or constraints without requiring their priorities to be expressed as weighted trade-offs~\citep{lesser2017multi, quan2025priority, lee2023hierarchical}. Hierarchical quadratic programming applies ordered equality and inequality constraints to humanoid motion generation~\citep{escande2014hierarchical}, while Rulebooks encode priorities among traffic rules for comparing driving behaviors~\citep{censi2019nutonomy}. More recently,~\citet{jia2026steering} combine gradient-based dynamic barriers with lexicographic selection to steer trajectories output by pretrained generative robot policies. TierCEM applies the same priority principle to the finite candidate set used in each CEM update, but incorporates it directly into the CEM optimization process to determine which candidates are retained and how the sampling distribution is updated. 

\paragraph{Planning and control with learned predictive models.}
Learned models supply predictions of future state given potential action sequences.
CEM is commonly used for reward-based planning with learned dynamics models~\citep{hafner2019learning, bharadhwaj2020model, kotb2023sample} and safety-aware planning~\citep{liu2020constrained, knuth2021planning}. Dreamer learns policies from imagined latent rollouts~\citep{hafner2019dream}, while SafeDreamer incorporates safety costs and Lagrangian-based planning into world-model control~\citep{huang2024safedreamer}. MUARL combines predicted safety costs and model uncertainty in sampling-based candidate selection through adaptive Lagrange multipliers~\citep{kolling2026evidential}. Within this literature, TierCEM targets planning problems with lexicographically-ordered constraints.
In our learned-model experiments, a frozen world model and learned constraint heads provide candidate evaluations, while TierCEM determines which requirements to preserve when no predicted candidate satisfies every constraint.

\section{Tiered cross-entropy method}
\label{sec:method}

TierCEM uses an ordered sequence of constraint filters to determine which candidates update CEM's sampling distribution. The method depends only on constraint evaluations and it is independent of the candidate representation. Algorithm~\ref{alg:lex_cem} summarizes the procedure, illustrated in  Figure~\ref{fig:tiercem}.

\subsection{Priority-relaxation cascade}
\label{sec:lex_cem}

\begin{algorithm}[!t]
  \caption{TierCEM}
  \label{alg:lex_cem}

  \algrenewcommand{\algorithmicindent}{1.25em}
  \algrenewcommand{\algorithmiccomment}[1]{%
    \hfill {\footnotesize\color{black!65}\itshape #1}%
  }

  \begin{algorithmic}
    \setlength{\itemsep}{0.15ex}

    \Require \begin{tabular}[t]{@{}l@{}}
      ordered margins $m_1 \succ \dots \succ m_N$,\\
      candidate samples $K$, elites $\ell$, iterations $I$
    \end{tabular}
    \smallskip

    \State Initialize $\mu, \Sigma$
    \For{$i = 1, \dots, I$}
    \State Sample $k$ candidates
    from $\mathcal{N}(\mu, \Sigma)$ and set \; $\mathcal{K}_0 \gets \{1,\dots,K\}$; \; $n \gets 1$
    \While{$n \leq N$}
    \State $\mathcal{K}_n \gets$ candidates in
    $\mathcal{K}_{n-1}$ satisfying constraint $n$
    \Comment{Equation~\eqref{eq:elites}}
    \If{$\mathcal{K}_n = \emptyset$} 
    \textbf{break}
    \Comment{retain $\mathcal{K}_{n-1}$} \EndIf
    \State $n \gets n + 1$
    \EndWhile
    \smallskip

    \State $r \gets \min(n, N)$
    \Comment{$n$ on fallback; $N$ otherwise}
    \State Compute $R(k) =  m_r(k)$
    for $k \in \mathcal{K}_{n-1}$
    \State $\mathcal{K}_{\mathrm{elite}}
    \gets \mathrm{Top}_{\ell}(\mathcal{K}_{n-1}, R)$
    \State Refit $\mu, \Sigma$ using the elite action sequences
    \EndFor
    \smallskip
    \State \Return $\mu, \Sigma$
  \end{algorithmic}
\end{algorithm}

\paragraph{Problem formulation.}
Let $\mathcal{X}$ denote the space of candidate solutions. We consider $N$ constraints, each represented by a margin function $m_n\colon\mathcal{X}\to\mathbb{R}$, where $m_n(x)>0$ indicates satisfaction of constraint $n$ and $m_n(x)\leq0$ indicates violation. The constraints are ordered by priority, $m_1 \succ m_2 \succ \dots \succ m_N$. Given a nonempty finite candidate set $\mathcal{K}\subset\mathcal{X}$, the goal is to select candidates according to the priority ordering; that is, to find candidates $k^*$ solving
\begin{align}
    k^* \in~ &\arg\max_{k \in \mathcal{K}} \quad m_{n^*}(k) \\
    &~\text{subject to} \quad m_n(k) > 0, ~ \text{for all } n < n^*
\end{align}
where $n^*$ is the highest priority constraint that \emph{cannot be satisfied together with all higher-priority constraints} (or $N$, the lowest-priority constraint, if all constraints can be satisfied simultaneously). 

\paragraph{Sampling.} At each iteration, $K$ candidates are sampled from the current Gaussian distribution $\mathcal{N}(\mu,\Sigma)$, forming the candidate set $\mathcal{K}$. TierCEM applies the ordered constraints to $\mathcal{K}$ to determine which candidates are used as elites for updating the sampling distribution.

\paragraph{Filtering.}
TierCEM enforces constraints from highest to lowest priority by
filtering the preceding candidate set. Starting from
$\mathcal{K}_0=\mathcal{K}$, tier $n$ retains 
\begin{equation}
  \mathcal{K}_n =
  \left\{
  k \in \mathcal{K}_{n-1}\;\vert\;
  m_n(k) > 0
  \right\}.
  \label{eq:elites}
\end{equation}
Because each filter acts on the preceding set, every candidate
in $\mathcal{K}_n$ satisfies constraints $1,\dots,n$.

\paragraph{Relaxation and ranking.}
In CEM, candidate samples are ranked according to a metric $R(k)$ (e.g. task reward), and the best $\ell$ candidate statistics, called elites, are used to update the Gaussian distribution. In TierCEM, if a tier empties the candidate set, its constraint governs elite 
ranking while the preceding constraints remain enforced. Suppose that 
$\mathcal{K}_{n^\ast}=\emptyset$ for some
$n^\ast$, and that $n^\ast$ is the first such tier. TierCEM returns to
$\mathcal{K}_{n^\ast-1}$ and ranks its candidates by 
\begin{equation}
  R(k) =
  m_{n^\ast}(k),
  \qquad k \in \mathcal{K}_{n^\ast-1}.
  \label{eq:tiercem-ranking}
\end{equation}

A larger score corresponds to a smaller worst-step violation of
constraint $n^\ast$. Every candidate in the retained set still
satisfies constraints $1,\dots,n^\ast-1$; lower-priority constraints
do not affect elite selection in that iteration. A candidate rejected
by an earlier filter therefore cannot re-enter elite selection
through better lower-priority margins.

If all filters retain candidates, TierCEM ranks $\mathcal{K}_N$
by $R(k)= m_N(k)$. The ranking then
favors a larger margin for the lowest-priority constraint
among candidates that satisfy every constraint.
In effect, this treats the function $m_N$ as an objective function which can only be optimized once higher priority constraints are satisfied.

The cascade provides a nonempty candidate set for ranking even
when no sampled trajectory satisfies every constraint. Relaxation
is determined by the prescribed ordering, without comparing
violation magnitudes across constraints or requiring the
penalty-weight calibration used in weighted control
formulations~\citep{wojsznis2007multi, qi2020enhancing}.

\paragraph{Elite selection and refitting.}
TierCEM selects the highest-ranked retained candidates as elites
and refits $\mathcal{N}(\mu,\Sigma)$ to their action sequences,
following the standard CEM update~\citep{botev2013cross}.
Sampling, filtering, ranking, and refitting are repeated for a fixed iteration budget of $I$ iterations. The best elite from the final iteration is returned, as done by~\citet{walker2026sample}. %

\section{Mechanism analysis under exact margins}
\label{sec:toy}

We conduct a series of empirical studies in the domain of trajectory optimization which isolate how TierCEM resolves constraint conflicts.
Specifically, we ask whether reversing the prescribed ordering changes which constraint is relaxed, and whether prioritizing task progress prevents the optimizer from preserving constraints by sacrificing progress. We study these mechanisms first in a controlled two-dimensional environment with closed-form dynamics (Section~\ref{sec:toy_progress}) and then in the contact-rich Push-T environment~\citep{chi2024diffusion} (Section~\ref{sec:pusht_proprio}). Both settings use exact, noise-free margin functions; Section~\ref{sec:pusht} then evaluates the same mechanism with learned margins in order to measure TierCEM's robustness to errors in learned constraint evaluations.

\subsection{States, dynamics, and constraints}
\label{sec:safe_wm}

We instantiate TierCEM in trajectory optimization. A discrete-time predictive model
$x_{t+1}=f(x_t,a_t)$ rolls the current state $x_0 \in \mathcal{X}$
forward under an action sequence $a_{0:H-1} \in \mathcal{A}^{H}$,
producing predicted states $x_{1:H}$. The margin function for each constraint $n \in \{1,\dots,N\}$ operates on the entire state-action trajectory. For example, a threshold $\bar a$ on the maximum control effort can be expressed as $m_n(x_{0:H-1}, a_{0:H-1}) = \min_t (\bar a - \|a_t\|)$, with $H$ being the planning horizon.

\subsection{Baselines}
\label{sec:softconstraint}

\paragraph{Scalarized CEM.} We compare TierCEM against two scalarized CEM baselines that rank
candidate trajectories using a weighted combination of task performance
and constraint margins:
\vspace{2em}
\begin{equation}
\label{eq:scalarized}
R^{(k)}
=
\eqnmarkbox[gray]{task}{\zeta\!\left(-\left\|p_H^{(k)} - p_{\mathrm{goal}}\right\|_2\right)}
+
\sum_{n=1}^{N}
\eqnmarkbox[orange]{weight}{\lambda_n}
\,
\zeta\!\left(\eqnmarkbox[blue]{margin}{m_n(x_t^{k})}
\right),
\end{equation}

\annotate[yshift=.5em]{above,left}{task}{task score}
\annotate[yshift=1em]{above,right}{weight}{manually tuned weight}
\annotate[yshift=-1em]{below,left}{margin}{constraint margin}
\vspace{0.4em}

\noindent
where $ -\left\|p_H^{(k)} - p_{\mathrm{goal}}\right\|_2$ is the negative terminal distance of the position $p_H$ of the task-relevant object
to the goal position $p_{\mathrm{goal}}$. The trajectory-level margin for constraint $n$ is defined as $m_n(x_{1:H}) = \frac{1}{H}\sum_{t=1}^{H}\tilde{m}_n(x_t)$, where $\tilde{m}_n(x_t)$ is the signed distance of state $x_t$ to the infeasible region for constraint $n$, and $\zeta(\cdot)$ standardizes each term over the candidate
population at the current CEM iteration. Soft-constraint CEM uses
a shared weight, $\lambda_n=\lambda$, whereas hierarchical
scalarized CEM uses separately tuned weights $\lambda_n$ to encode
the prescribed priorities. Setting $\lambda_n=0$ for all $n$
recovers task-only CEM.

\paragraph{Constrained cross-entropy.} 
We also compare against CCE~\citep{wen2018constrained}, 
which separates feasibility from task optimization 
rather than combining them through a weighted objective. 
When too few sampled policies are feasible, CCE updates its sampling distribution according 
to constraint violation, once sufficient feasible policies are found,
it selects according to task performance. 
CCE therefore provides a feasibility-first comparison, 
but treats the constraints jointly rather than specifying which constraint should be relaxed when they conflict.

\subsection{Priority-controlled relaxation in 2D navigation}
\label{sec:toy_progress}

We construct a two-dimensional navigation problem in which reaching the goal requires violating one of two spatial constraints. A lane connects the start and goal but intersects a forbidden region: remaining inside the lane requires crossing the forbidden region, whereas avoiding the forbidden region requires leaving the lane. The system has single-integrator dynamics, so a zero action can maintain both spatial constraints indefinitely without making progress toward the goal. 

\begin{figure}[!t]
    \centering
    \def\svgwidth{\linewidth}
    \begingroup%
  \makeatletter%
  \providecommand\color[2][]{%
    \errmessage{(Inkscape) Color is used for the text in Inkscape, but the package 'color.sty' is not loaded}%
    \renewcommand\color[2][]{}%
  }%
  \providecommand\transparent[1]{%
    \errmessage{(Inkscape) Transparency is used (non-zero) for the text in Inkscape, but the package 'transparent.sty' is not loaded}%
    \renewcommand\transparent[1]{}%
  }%
  \providecommand\rotatebox[2]{#2}%
  \newcommand*\fsize{\dimexpr\f@size pt\relax}%
  \newcommand*\lineheight[1]{\fontsize{\fsize}{#1\fsize}\selectfont}%
  \ifx\svgwidth\undefined%
    \setlength{\unitlength}{1567.36043615bp}%
    \ifx\svgscale\undefined%
      \relax%
    \else%
      \setlength{\unitlength}{\unitlength * \real{\svgscale}}%
    \fi%
  \else%
    \setlength{\unitlength}{\svgwidth}%
  \fi%
  \global\let\svgwidth\undefined%
  \global\let\svgscale\undefined%
  \makeatother%
  \begin{picture}(1,0.28367147)%
    \lineheight{1}%
    \setlength\tabcolsep{0pt}%
    \put(0,0){\includegraphics[width=\unitlength,page=1]{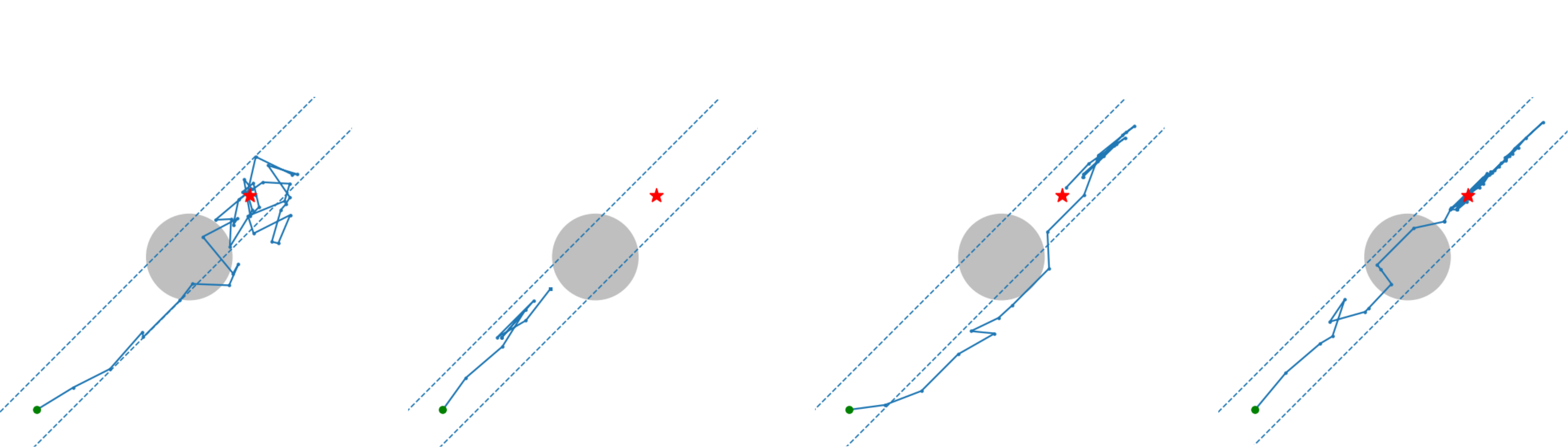}}%
    \put(0.09519325,0.27034319){\color[rgb]{0,0,0}\makebox(0,0)[lt]{\lineheight{1.25}\smash{\begin{tabular}[t]{l}\textbf{CEM}\end{tabular}}}}%
    \put(0.36459964,0.27203407){\color[rgb]{0,0,0}\makebox(0,0)[t]{\lineheight{1.25}\smash{\begin{tabular}[t]{c}\textbf{task last}\end{tabular}}}}%
    \put(0.64446033,0.27137363){\color[rgb]{0,0,0}\makebox(0,0)[t]{\lineheight{1.25}\smash{\begin{tabular}[t]{c}\textbf{obstacle $\succ$ lane}\end{tabular}}}}%
    \put(0.88102679,0.27091412){\color[rgb]{0,0,0}\makebox(0,0)[t]{\lineheight{1.25}\smash{\begin{tabular}[t]{c}\textbf{lane $\succ$ obstacle}\end{tabular}}}}%
  \end{picture}%
\endgroup%

    \caption{Effect of the prescribed priority ordering in the two-dimensional navigation problem. When task progress has lower priority than the spatial constraints, TierCEM can preserve both by stalling. Prioritizing task progress forces a spatial constraint to relax; reversing the spatial ordering reverses whether the trajectory leaves the lane or crosses the forbidden region.}
    \label{fig:toy_main}
\end{figure}

We use this construction to separate the effect of task priority from the effect of the ordering between spatial constraints. In \emph{task\_last}, the lane and forbidden-region constraints precede task progress, allowing TierCEM to preserve both spatial constraints by stalling (Figure~\ref{fig:toy_main}). The two task-first configurations instead require progress and reverse the spatial priorities: \emph{task\_obstacle\_lane} prioritizes obstacle avoidance over lane keeping, whereas \emph{task\_lane\_obstacle} does the opposite. All constraint margins are exact signed distances. Planner hyperparameters are reported in Appendix~\ref{apx:toy_setup}, Table~\ref{tab:appx_toy_config}.

\begin{table}[b]
\centering
\caption{TierCEM versus the Soft-Constraint CEM, Constrained Cross-Entropy~\citep{wen2018constrained} (Section~\ref{sec:softconstraint}), and the hierarchical scalarized CEM (Section~\ref{sec:softconstraint}) on the task described in Section~\ref{sec:toy_progress} across 20 random seeds, using 10 CEM iterations, initial Gaussian variance per dimension 1.00, planning horizon $H = 8$, 4 committed actions before re-planning, 1200 samples and 24 elites per iteration. ``Reach" shows the number of seeds where the method reaches the goal out of the 20 different seeds,  \emph{V$_\uparrow$}/\emph{V$_\downarrow$} count seeds where the higher-/lower-priority constraint are violated and ``Steps'' reflects the average number of steps required to complete the task. Hierarchical scalarized CEM shows the weights as tuples ($\lambda_O$, $\lambda_L$) for each constraint priority order.}
\label{tab:toy_softbaseline}
\resizebox{\linewidth}{!}{%
\begin{tabular}{lcccclcccc}
\toprule
 & \multicolumn{4}{c}{\tt{task\_obstacle\_lane}} & & \multicolumn{4}{c}{\tt{task\_lane\_obstacle}} \\
\textbf{Method} & \textbf{Reach} & \textbf{V$_\uparrow$} & \textbf{V$_\downarrow$} & \textbf{Steps} & & \textbf{Reach} & \textbf{V$_\uparrow$} & \textbf{V$_\downarrow$} & \textbf{Steps} \\
\midrule
TierCEM   & 20/20 & 0 & 20 & 34 & & 20/20 & 0 & 20 & 32 \\
CCE~\citep{wen2018constrained} & 0/20 & 0 & 0 & 160 & & 0/20 & 0 & 0 & 160 \\
Soft-constraint CEM, $\lambda = 0.5$ & 20/20 & 20 & 12 & 30 & & 20/20 & 11 & 20 & 21 \\
Hierarchical scalarized CEM, (0.5, 0.02)/(0.02, 0.5) & 20/20 & 0 & 20 & 38 & & 20/20 & 0 & 20 & 16 \\

\bottomrule
\end{tabular}}
\end{table}

Prioritizing task progress converts constraint-preserving stalling into controlled constraint relaxation. CCE preserves both spatial constraints in every episode but never reaches the goal. In contrast, soft-constraint CEM reaches the goal in every episode but frequently violates the higher-priority constraint, since its shared weight does not distinguish the two spatial priorities. 

Table~\ref{tab:toy_softbaseline} shows that
TierCEM and hierarchical scalarized CEM both reach the goal while preserving the higher-priority spatial constraint in all 40 episodes. Their difference lies in how the priority is specified. Hierarchical scalarization requires reversing the numerical constraint weights when the priority ordering is reversed, whereas TierCEM changes only the ordering of its constraint filters. For TierCEM, reversing that ordering reverses which lower-priority constraint is violated (Figure~\ref{fig:toy_main}). 

The stalling mechanism depends on dynamics that admit a stationary constraint-satisfying solution; it does not arise in the inertial setting examined in Appendix~\ref{apx:toy_inertial}. We additionally repeat the ordering intervention under alternative constraint geometries in Appendix~\ref{apx:toy_ordering}.

\subsection{Contact-rich manipulation with exact margins}
\label{sec:pusht_proprio}

\begin{figure}[t]
    \centering
    \def\svgwidth{0.8\linewidth}
    \begingroup%
  \makeatletter%
  \providecommand\color[2][]{%
    \errmessage{(Inkscape) Color is used for the text in Inkscape, but the package 'color.sty' is not loaded}%
    \renewcommand\color[2][]{}%
  }%
  \providecommand\transparent[1]{%
    \errmessage{(Inkscape) Transparency is used (non-zero) for the text in Inkscape, but the package 'transparent.sty' is not loaded}%
    \renewcommand\transparent[1]{}%
  }%
  \providecommand\rotatebox[2]{#2}%
  \newcommand*\fsize{\dimexpr\f@size pt\relax}%
  \newcommand*\lineheight[1]{\fontsize{\fsize}{#1\fsize}\selectfont}%
  \ifx\svgwidth\undefined%
    \setlength{\unitlength}{1498.14120555bp}%
    \ifx\svgscale\undefined%
      \relax%
    \else%
      \setlength{\unitlength}{\unitlength * \real{\svgscale}}%
    \fi%
  \else%
    \setlength{\unitlength}{\svgwidth}%
  \fi%
  \global\let\svgwidth\undefined%
  \global\let\svgscale\undefined%
  \makeatother%
  \begin{picture}(1,0.19215957)%
    \lineheight{1}%
    \setlength\tabcolsep{0pt}%
    \put(0,0){\includegraphics[width=\unitlength,page=1]{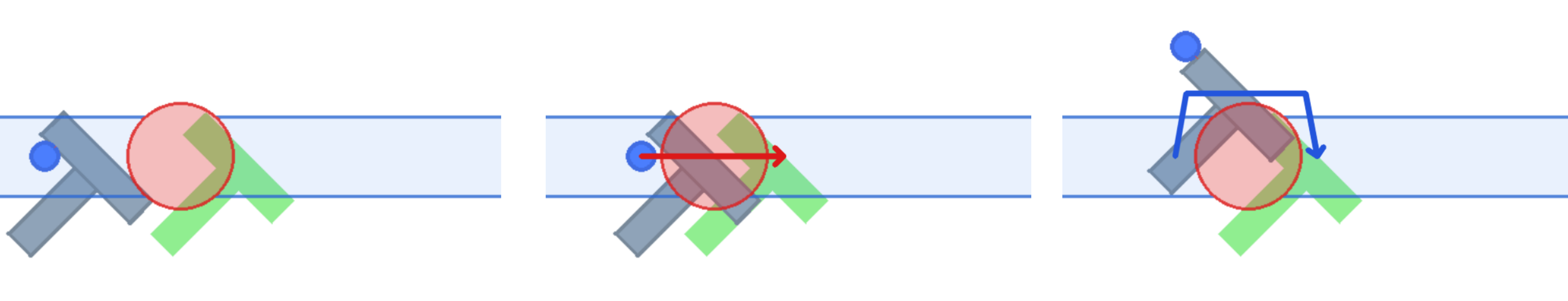}}%
    \put(0.17070259,0.17759728){\color[rgb]{0,0,0}\makebox(0,0)[t]{\lineheight{1.25}\smash{\begin{tabular}[t]{c}\textbf{Push-T scenario}\end{tabular}}}}%
    \put(0.49715919,0.17617628){\color[rgb]{0,0,0}\makebox(0,0)[t]{\lineheight{1.25}\smash{\begin{tabular}[t]{c}\textbf{lane $\succ$ obstacle}\end{tabular}}}}%
    \put(0.84793718,0.17795535){\color[rgb]{0,0,0}\makebox(0,0)[t]{\lineheight{1.25}\smash{\begin{tabular}[t]{c}\textbf{obstacle $\succ$ lane}\end{tabular}}}}%
  \end{picture}%
\endgroup%

    \caption{Visualization of the competing constraints used in the Push-T environment. The goal is to move the block toward the goal position, ignoring orientation. First image shows the initial position, the second image shows the expected behavior when staying within the lane boundary is prioritized over the obstacle constraint, and the third image shows the expected behavior when the obstacle constraint is prioritized over staying within the lane boundary.
    }
    \label{fig:pusht_conflict}
\end{figure}

We next test whether the same priority-controlled relaxation persists under  contact-rich dynamics, which make the trajectory optimization problems far harder to solve. We transfer the lane-obstacle conflict to the Push-T environment~\citep{chi2024diffusion}, where a circular end-effector must push the block toward its goal while navigating a narrow region defined by a lane and a forbidden zone (cf. Figure~\ref{fig:pusht_conflict}). The geometry permits only a narrow set of trajectories that satisfy both spatial constraints while completing the task. We retain exact simulator state and constraint margins, changing the dynamics while preserving the mechanism isolated in the two-dimensional environment.

Reversing the TierCEM ordering reverses the violation pattern while leaving task performance similar. Prioritizing obstacle avoidance nearly eliminates obstacle violations while allowing the end-effector to leave the lane. In fact, prioritizing lane keeping substantially reduces lane violations while permitting obstacle violations (Table~\ref{tab:pusht_oracle_conflict}). Hierarchical scalarized CEM produces the same qualitative ordering-dependent behavior, but requires tuning the numerical constraint weights, a difficult process especially as the number of constraints grows and if the scaling is different among constraints. Soft-constraint CEM, which uses a shared weight, does not encode which spatial constraint should be preserved. 

At the nominal goal (that is, the green region of Figure~\ref{fig:pusht_conflict}), CCE satisfies both spatial constraints with task error comparable to TierCEM, so this setting alone does not require constraint relaxation. We therefore increase the required task progress by moving the goal farther from the initial block position while keeping planner hyperparameters fixed.

\begin{table}[t]
\centering
\caption{Exact-state (proprioceptive) evaluation on Push-T.
Values are mean $\pm$ standard deviation across 20 random seeds. Position error is the final Euclidean block-center distance to the goal, violation rates are fractions of executed environment steps. Although CCE can converge to feasible solutions with a large number of candidate samples (1200 in the results of this table), its feasibility-first objective can prevent task completion under limited sampling when jointly feasible trajectories are rare. For example, with 300 samples, CCE often fails to discover task-completing trajectories and instead prioritizes constraint satisfaction (Table~\ref{tab:pusht_wm_conflict}). Instead, TierCEM can relax the lower-priority constraint and retain task completion.}
\label{tab:pusht_oracle_conflict}
\begin{adjustbox}{width=\linewidth}
\begin{tabular}{llccc}
\toprule
\textbf{Method} & \textbf{Config} & \textbf{Pos. error (px)} & \textbf{O viol. (\%)} & \textbf{Lane viol. (\%)} \\
\midrule
TierCEM & task $\succ$ obstacle $\succ$ lane & $20.63 \pm 16.17$ & $0.5 \pm 1.5$ & $64.8 \pm 34.3$ \\
TierCEM & task $\succ$ lane $\succ$ obstacle & $22.34 \pm 12.23$ & $44.0 \pm 15.9$ & $10.3 \pm 9.8$ \\
CCE~\citep{wen2018constrained} & joint worst violation & $21.29 \pm 18.13$ & $0.4 \pm 1.2$ & $0.2 \pm 1.1$ \\
Soft-constraint CEM &  $\lambda=0.5$ & $17.30 \pm 11.99$ & $19.5 \pm 13.3$ & $15.3 \pm 11.3$ \\
Hierarchical scalarized CEM & $\lambda_{O} = 0.5, \lambda_{L} = 0.02$ & $15.00 \pm 13.48$ & $1.2 \pm 3.4$ & $71.7 \pm 37.1$ \\
Hierarchical scalarized CEM & $\lambda_{O} = 0.02, \lambda_{L} =  0.5$ & $18.09 \pm 10.26$ & $29.9 \pm 13.7$ & $11.7 \pm 10.9$ \\
\bottomrule
\end{tabular}

\end{adjustbox}
\end{table}

\begin{table}[t]
\centering
\caption{Goal-distance sensitivity in exact-state Push-T, with all planner
hyperparameters fixed. Values are mean $\pm$ standard deviation across 20
random seeds. Position error is measured in pixels; obstacle and lane
violations are percentages of executed controls. Increasing goal distance
turns jointly safe satisfaction into lower task accuracy for CCE, while
TierCEM can continue by relaxing its designated lower-priority lane
constraint.}
\label{tab:pusht_oracle_goal_distance}
\resizebox{\linewidth}{!}{%
\begin{tabular}{llccc}
\toprule
\textbf{Method/configuration}
    & \textbf{Metric}
    & \textbf{Goal $x=256$}
    & \textbf{Goal $x=320$}
    & \textbf{Goal $x=360$} \\
\midrule
TierCEM, task $\succ$ obstacle $\succ$ lane
    & Position error (px)
    & $20.63 \pm 16.17$
    & $16.06 \pm 19.06$
    & $30.96 \pm 34.31$ \\
    & Obstacle violation (\%)
    & $0.50 \pm 1.50$
    & $3.75 \pm 6.78$
    & $3.08 \pm 6.71$ \\
    & Lane violation (\%)
    & $64.75 \pm 34.26$
    & $71.58 \pm 32.98$
    & $57.25 \pm 33.70$ \\
\midrule
CCE~\citep{wen2018constrained}, joint constraints
    & Position error (px)
    & $21.29 \pm 18.13$
    & $56.35 \pm 30.77$
    & $87.80 \pm 36.79$ \\
    & Obstacle violation (\%)
    & $0.38 \pm 1.19$
    & $1.51 \pm 2.32$
    & $3.45 \pm 4.56$ \\
    & Lane violation (\%)
    & $0.25 \pm 1.09$
    & $1.57 \pm 2.59$
    & $3.34 \pm 3.44$ \\
\bottomrule
\end{tabular}}
\end{table}
Increasing the required progress separates feasibility-first selection from ordered relaxation. CCE's final position error increases as the goal moves farther away, whereas TierCEM continues toward the goal by relaxing its designated lower-priority constraint (Table~\ref{tab:pusht_oracle_goal_distance}). Together with the two-dimensional experiment, these results show that the prescribed ordering controls which requirement is relaxed under both kinematic and contact-rich dynamics when constraint evaluations are exact.

\section{Planning with learned world-model predictions}
\label{sec:pusht}

The experiments of Section~\ref{sec:pusht_proprio} isolate the optimizer from representation and dynamics errors. We now evaluate the same conflictive constraint scenario using learned margins over DINO-WM latents~\citep{pmlr-v267-zhou25t} with the same CEM hyperparameter configuration as~\citet{terver2026drives}. This comparison tests whether the priority-relaxation behavior survives when candidate trajectories, task progress, and constraint margins are inferred from predicted visual latents rather than exact simulator states. TierCEM remains effective when the constraints are jointly satisfiable (Appendix~\ref{apx:noconflict}). Under imperfect constraint and dynamics models, errors in the margin function prediction can lead to conservative behavior, whereas dynamics mismatch can result in constraint violations (Appendix~\ref{apx:imperfect}).%

The environment and the constraints are identical to Section~\ref{sec:pusht_proprio}. Each learned margin function is implemented as a multi-layer perceptron (MLP) that outputs the probability of violation, which we expose to TierCEM as the margin $m_n(z_t)=\min_{t=1,\ldots,H} [0.5-\mathbb{P}_n(\mathrm{violation}\mid z_t)]$, where $z_t$ is the latent representation of $x_t$.
Additionally, we train an MLP  regressor to predict negative distance to goal, $y_t = -\|p_{\mathrm{block},t} - p_\mathrm{goal}\|_2$. We define the task progress margin for a candidate trajectory $k$ as the reduction in distance over the horizon: $m_\mathrm{task}^{(k)}=y_H^{(k)}-y_0^{(k)}$. Positive values indicate progress toward the goal. 

\paragraph{Learned margin function results.} Using the MLP hyperparameters (Appendixes~\ref{apx:mlp_hyper} and~\ref{apx:task_progress_mlp}), the MLP predicting $\mathbb{P}_n(\mathrm{violation}\mid z_t)$ is capable of identifying violating and non-violating latents for the obstacle and the lane constraints, achieving strong separability of both classes for both constraints. The MLP regressor also achieves a small error in predicting the distance to the goal, with a mean absolute error of 6.2 px. Figure~\ref{fig:pusht_mlp} shows a visual representation of the results obtained. %

\begin{figure*}[t]
    \centering
    \begin{subfigure}[t]{\textwidth}
        \centering
        \includegraphics[width=\linewidth]{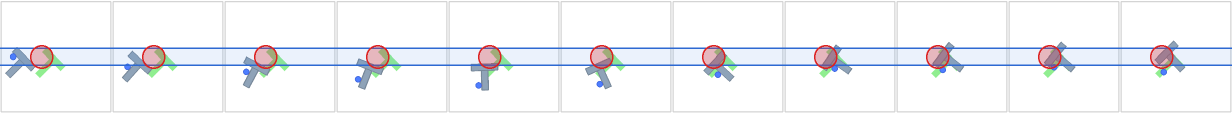}
        \caption{TierCEM, priority order: task $\succ$ obstacle $\succ$ lane.}
    \end{subfigure}
    \hfill
    \begin{subfigure}[t]{\textwidth}
        \centering
        \includegraphics[width=\linewidth]{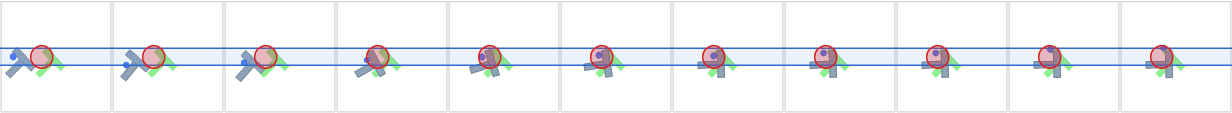}
        \caption{TierCEM, priority order: task $\succ$ lane $\succ$ obstacle.}
    \end{subfigure}
    \hfill
    \begin{subfigure}[t]{\textwidth}
        \centering
        \includegraphics[width=\linewidth]{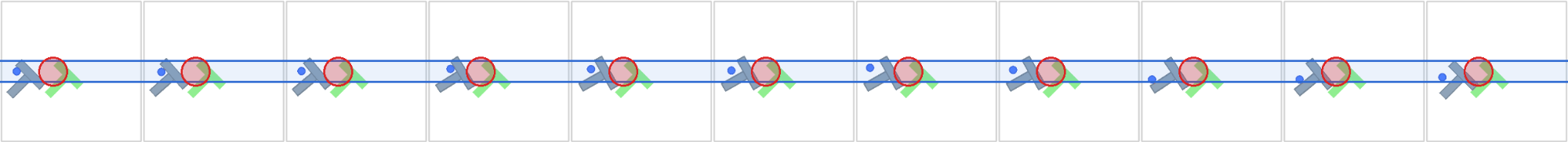}
        \caption{Constrained Cross Entropy (CCE)~\citep{wen2018constrained}}
    \end{subfigure}
    \hfill
    \begin{subfigure}[t]{\textwidth}
        \centering
        \includegraphics[width=\linewidth]{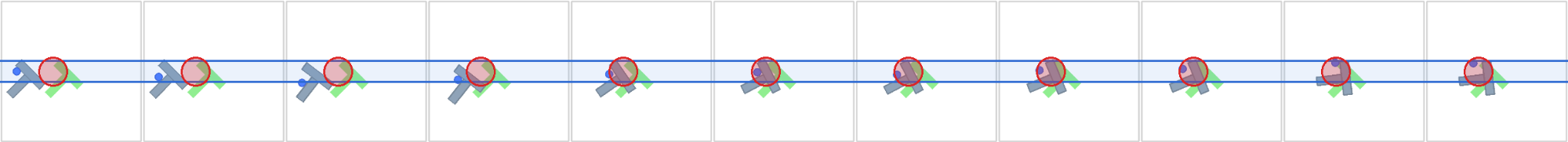}
        \caption{Soft-Constraint CEM, $\lambda = 0.5$}
    \end{subfigure}
    \hfill
    \begin{subfigure}[t]{\textwidth}
        \centering
        \includegraphics[width=\linewidth]{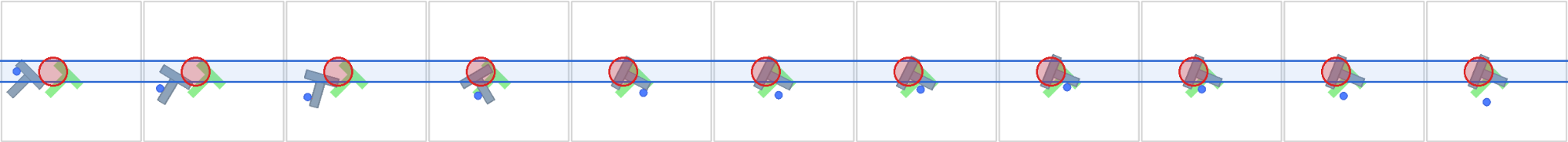}
        \caption{Hierarchical scalarized CEM, $\lambda_{O}=0.5, \lambda_L = 0.02$.}
    \end{subfigure}
    \hfill
    \begin{subfigure}[t]{\textwidth}
        \centering
        \includegraphics[width=\linewidth]{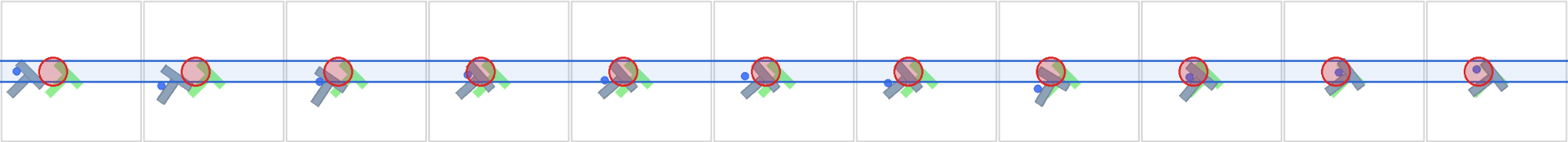}
        \caption{Hierarchical scalarized CEM, $\lambda_{O}=0.02, \lambda_L = 0.5$.}
    \end{subfigure}
    \caption{TierCEM and baselines execution in the Push-T scenario using DINO-WM latents. Frames progress from left to right. The red disk blocks the blue lane, the green T region is the goal, and the blue T is the block. TierCEM relaxes the lower-priority constraint to complete the task without requiring a per-constraint weight tuning as hierarchical CEM, CCE preserves safety without progressing toward the goal, and the Soft-Constraint CEM does not have any mechanism to prioritize one constraint over the other. %
    }
    \label{fig:pusht_conflict_wm}
\end{figure*}

\begin{table}[t]
\centering
\caption{TierCEM and baseline methods on the Push-T scenario using DINO-WM latents. Values are mean $\pm$ standard deviation across 20 episodes per configuration. Position error is the final Euclidean block-center distance to the goal. O/Lane violation is the percentage of timesteps in violation.}
\label{tab:pusht_wm_conflict}
\begin{adjustbox}{width=\linewidth}
\begin{tabular}{llccc}
\toprule
\textbf{Method} & \textbf{Config} & \textbf{Pos. error (px)} & \textbf{O viol. (\%)} & \textbf{Lane viol. (\%)} \\
\midrule
TierCEM & task $\succ$ obstacle $\succ$ lane & $57.95 \pm 11.09$ & $10.4 \pm17.4$ & $66.4 \pm  27.3$ \\
TierCEM & task $\succ$ lane $\succ$ obstacle & $45.32 \pm 7.63$ & $54.3 \pm 20.9$ & $3.2 \pm 7.7$ \\
CCE~\citep{wen2018constrained} & joint worst violation & $105.11 \pm 17.54$ & $0.4 \pm 1.2$ & $1.6 \pm 3.1$ \\
Soft-Constraint CEM & $\lambda = 0.5$ & $21.61 \pm  6.26$ & $34.8 \pm 11.9$  & $19.9 \pm 5.1$ \\
Hierarchical scalarized CEM & $\lambda_{O} = 0.5, \lambda_{L} = 0.02$ & $46.49 \pm 22.40$ & $12.0  \pm 10.8$ & $79.2 \pm 16.5$ \\
Hierarchical scalarized CEM & $\lambda_{O} = 0.02, \lambda_{L} = 0.5$ & $14.93 \pm 12.21$ & $47.4 \pm 13.3$ & $26.4 \pm 12.3$ \\
Hierarchical scalarized CEM & $\lambda_{O} = 0.5, \lambda_{L} = 0.1$ & $27.80  \pm 15.61$ & $30.0  \pm 11.9$ & $51.8 \pm 13.85$ \\
Hierarchical scalarized CEM & $\lambda_{O} = 0.1, \lambda_{L} = 0.5$ & $13.70   \pm 8.17$ & $37.8  \pm 19.5$ & $27.8 \pm 11.7$ \\
\bottomrule
\end{tabular}
\end{adjustbox}
\end{table}

\paragraph{TierCEM behavior.} TierCEM preserves the behavior of Section~\ref{sec:pusht_proprio} despite errors in the learned dynamics and constraint margins. Under the configuration task $\succ$ obstacle $\succ$ lane , the end-effector goes around the obstacle and TierCEM relaxes the lane, while under task $\succ$ lane $\succ$ obstacle, the end-effector stays in the lane boundary and crosses the obstacle. This means, the higher-priority constraint remains better satisfied while the incompatible lower-priority constraint absorbs the violation. Figure~\ref{fig:pusht_conflict_wm} shows an example trajectory generated by TierCEM and each of the baselines.

The results of Table~\ref{tab:pusht_wm_conflict} show that the behavior of hierarchical scalarization is sensitive to the choice of penalty weights. For example, under task $\succ$ obstacle $\succ$ lane, with $\lambda_{\mathrm{O}}=0.5$ fixed, decreasing $\lambda_{\mathrm{L}}$ from $0.1$ to $0.02$ increases lane violations, while obstacle violations decrease. TierCEM instead only requires the priority order, and obtains lower obstacle and lane violations for the same priority configuration.

\begin{remark}
The value of the weights of the hierarchical scalarized CEM do not transfer from task to task in the same environment. Table~\ref{tab:pusht_wm_conflict} shows that $\lambda_\mathrm{O} = 0.5, \lambda_{L} = 0.02$ and vice versa obtain the best results in terms of high-priority constraint satisfaction. However, $\lambda_\mathrm{O} = 0.5, \lambda_{L} = 0.1$ and vice versa is the best configuration according to Table~\ref{tab:hier-scalarized} for a different Push-T variant with a non-conflictive constraint setup (Appendix~\ref{apx:noconflict}).
\end{remark}

\section{Conclusion}

When all constraints cannot be satisfied simultaneously, optimizers must determine which constraints to relax. TierCEM addresses this by encoding the constraint priorities as an explicit lexicographic ordering, rather than requiring the relative importance of constraints to be expressed through weights. Our experiments show that, when constraints conflict, TierCEM relaxes the lower-priority constraint, and reversing the priority order changes which constraint is relaxed. We also show that a feasibility-oriented optimizer without an explicit mechanism for choosing which constraint to relax, such as CCE~\citep{wen2018constrained}, are designed to sacrifice task completion in favor of constraint satisfaction. This behavior is consistent with their objective, but can produce constraint-satisfying yet task-irrelevant candidate samples. Our problem formulation allows us to express task progress as a constraint that the optimizer can prioritize, allowing task completion to be retained when constraints conflict.  We show that TierCEM can function effectively even when used with learned constraint specifications and in highly complex, contact-rich trajectory optimization problems; scenarios in which TierCEM's performance depends on the accuracy of the learned dynamics and constraint models, meaning that deployment in safety-critical applications would require appropriate safety guarantees.

\section*{Acknowledgments}

F. Roldan Sanchez, P. de las Heras Molins, and G. Bakirtzis are partially supported by the academic and research chair Architecture des Systèmes Complexes through the following partners: Dassault Aviation, Naval Group, Dassault Systèmes, KNDS France, Agence de l’Innovation de Défense, and Institut Polytechnique de Paris.
D. Fridovich-Keil is partially supported by the Hi! PARIS international visiting chair program.

\FloatBarrier

\bibliography{typescript}

@inproceedings{barkley2025scoped,
  title={{SCOPED}: Score-Curvature Out-of-distribution Proximity Evaluator for Diffusion},
  author={Barkley, B. and Culbertson, P. and Fridovich-Keil, D.},
  booktitle={ICLR},
  year={2025},
url = {https://openreview.net/forum?id=TMLiG9Rk2J}
}

@article{wen2018constrained,
  author={Wen, M. and Topcu, U.},
  journal={Transactions on Automatic Control}, 
  title={Constrained Cross-Entropy Method for Safe Reinforcement Learning}, 
  year={2021},
  doi={10.1109/TAC.2020.3015931}
  }

@article{liu2024dynamic,
  title={Dynamic $\varepsilon$-multilevel hierarchy constraint optimization with adaptive boundary constraint handling technology},
  author={Liu, J. and Feng, J. and Yang, S. and Zhang, H. and Liu, S.},
  journal={Appl. Soft Comput.},
  year={2024},
  doi={10.1016/j.asoc.2023.111172}
}

@inproceedings{hafner2019learning,
  title={Learning latent dynamics for planning from pixels},
  author={Hafner, D. and Lillicrap, T. and Fischer, I. and Villegas, R. and Ha, D. and Lee, H. and Davidson, J.},
  booktitle={ICML},
  year = 	 {2019},
  url = 	 {https://proceedings.mlr.press/v97/hafner19a.html},
}

@inproceedings{kotb2023sample,
  title={Sample-efficient real-time planning with curiosity cross-entropy method and contrastive learning},
  author={Kotb, M. and Weber, C. and Wermter, S.},
  booktitle={IROS},
  year={2023},
  doi={10.1109/IROS55552.2023.10342018}
}

@inproceedings{bharadhwaj2020model,
  title={Model-predictive control via cross-entropy and gradient-based optimization},
  author={Bharadhwaj, H. and Xie, K. and Shkurti, F.},
  booktitle = 	 {L4DC},
  year = 	 {2020},
  url = 	 {https://proceedings.mlr.press/v120/bharadhwaj20a.html},
}

@article{liu2020constrained,
  title={Constrained model-based reinforcement learning with robust cross-entropy method},
  author={Liu, Z. and Zhou, H. and Chen, B. and Zhong, S. and Hebert, M. and Zhao, D.},
  journal={arXiv:2010.07968},
  year={2020},
  doi={10.48550/arXiv.2010.07968}
}

@inproceedings{huang2024safedreamer,
  title={{SafeDreamer}: Safe reinforcement learning with world models},
  author={Huang, W. and Ji, J. and Xia, C. and Zhang, B. and Yang, Y.},
  booktitle={ICLR},
  year={2024},
  url          = {https://openreview.net/forum?id=tsE5HLYtYg},
}

@inproceedings{hafner2019dream,
  title={Dream to control: Learning behaviors by latent imagination},
  author={Hafner, D. and Lillicrap, T. and Ba, J. and Norouzi, M.},
  booktitle={ICLR},
  url          = {https://openreview.net/forum?id=S1lOTC4tDS},
  year={2020}
}

@article{knuth2021planning,
  title={Planning with learned dynamics: Probabilistic guarantees on safety and reachability via {L}ipschitz constants},
  author={Knuth, C. and Chou, G. and Ozay, N. and Berenson, D.},
  journal={RA-L},
  year={2021},
  publisher={IEEE},
doi = {10.1109/LRA.2021.3068889}
}

@inproceedings{kolling2026evidential,
  title={Evidential Latent World Models for Safe Model-Based Reinforcement Learning},
  author={Kolling, A. H. and De Jesus, J. C. and Kich, V. A. and Grando, R. B. and Mateus, M. G. and da Silva Guerra, R. and Drews-Jr, P. L. J.},
  booktitle={ICLR Workshop on World Models},
  year={2026},
url = {https://openreview.net/pdf?id=8P1pUpVKvA}
}

@inproceedings{mottahedi2022constrained,
  title={Constrained differentiable cross-entropy method for safe model-based reinforcement learning},
  author={Mottahedi, S. and Pavlak, G.},
  booktitle={BuildSys},
  year={2022},
  doi={10.1145/3563357.3564055}
}

@article{chi2024diffusion,
  title={Diffusion policy: Visuomotor policy learning via action diffusion},
  author={Chi, C. and Xu, Z. and Feng, S. and Cousineau, E. and Du, Y. and Burchfiel, B. and Tedrake, R. and Song, S.},
  journal={Int. J. Robot. Res.},
  year={2024},
  doi={10.1177/02783649241273668}
}

@article{wojsznis2007multi,
  title={Multi-objective optimization for model predictive control},
  author={Wojsznis, W. and Mehta, A. and Wojsznis, P. and Thiele, D. and Blevins, T.},
  journal={ISA Trans.},
  year={2007},
  doi={10.1016/j.isatra.2006.10.002}
}

@article{qi2020enhancing,
  title={Enhancing dynamic operation optimization feasibility for constrained model predictive control systems},
  author={Qi, X. and Li, S. and Zheng, Y.},
  journal={ASJC},
  year={2020},
  doi={10.1002/asjc.2158}
}

@InProceedings{pmlr-v267-zhou25t,
  title ={{DINO-WM}: World Models on Pre-trained Visual Features enable Zero-shot Planning},
  author = {Zhou, G. and Pan, H. and Lecun, Y. and Pinto, L.},
  booktitle = {ICML},
  year={2025},
  url = {https://proceedings.mlr.press/v267/zhou25t.html}
}

@article{terver2026drives,
  title={What Drives Success in Physical Planning with Joint-Embedding Predictive World Models?},
  author={Terver, B. and Yang, T. and Ponce, J. and Bardes, A. and LeCun, Y.},
  journal={TMLR},
  year={2026},
url={https://openreview.net/forum?id=cHZn5Gdh8e}
}

@article{quan2025priority,
  title={Priority-driven constraints softening in safe mpc for perturbed systems},
  author={Quan, Y. S. and Jeddi, M. and Prignoli, F. and Falcone, P.},
  journal={L-CSS},
  year={2025},
  doi={10.1109/LCSYS.2025.3580494}
}

@inproceedings{lesser2017multi,
  title={Multi-objective optimal control with safety as a priority},
  author={Lesser, K. and Abate, A.},
  booktitle={ICCPS},
  year={2017},
  doi={10.1145/3055004.3055008}
}

@inproceedings{lee2023hierarchical,
  title={Hierarchical relaxation of safety-critical controllers: Mitigating contradictory safety conditions with application to quadruped robots},
  author={Lee, J. and Kim, J. and Ames, A.},
  booktitle={IROS},
  year={2023},
  doi={10.48550/arXiv.2305.03929}
}

@article{botev2013cross,
  title={The cross-entropy method for optimization},
  author={Botev, Z. and Kroese, D. and Rubinstein, R. and  L'Ecuyer, P.},
  journal={Handbook of Statistics},
  year={2013},
  doi={10.1016/B978-0-444-53859-8.00003-5}
}

@inproceedings{censi2019nutonomy,
  author={Censi, A. and Slutsky, K. and Wongpiromsarn, T. and Yershov, D. and Pendleton, S. and Fu, J. and Frazzoli, E.},
  booktitle={ICRA}, 
  title={Liability, Ethics, and Culture-Aware Behavior Specification using Rulebooks}, 
  year={2019},
  doi={10.1109/ICRA.2019.8794364}
  }

@article{williams2017model,
  title={Model predictive path integral control: From theory to parallel computation},
  author={Williams, G. and Aldrich, A. and Theodorou, E.},
  journal={Guidance, Control, and Dynamics},
  year={2017},
  doi={10.2514/1.G001921}
}

@article{imanov2026autonomous,
  title={Autonomous {AI} Agents for Real-Time Affordable Housing Site Selection: Multi-Objective Reinforcement Learning Under Regulatory Constraints},
  author={Imanov, O. and Erisken, D. and Kulali, D. and Yilmaz, T. and Turhan, R.},
  journal={arXiv:2602.03940 },
  year={2026}, 
  doi={10.48550/arXiv.2602.03940}
}

@article{escande2014hierarchical,
  title={Hierarchical quadratic programming: Fast online humanoid-robot motion generation},
  author={Escande, A. and Mansard, N. and Wieber, P.},
  journal={Int. J. Robot. Res.},
  year={2014},
  doi={10.1177/0278364914521306}
}

@article{jia2026steering,
  title={Steering Generative Robot Policies with Lexicographic Preferences},
  author={Jia, Y. and How, J.},
  journal={arXiv:2609.15014 },
  year={2026},
  doi={10.48550/arXiv.2609.15014}
}

@inproceedings{walker2026sample,
  title={Sample-efficient and smooth cross-entropy method model predictive control using deterministic samples},
  author={Walker, M. and Frisch, D. and Hanebeck, U.},
  booktitle={ACC},
  year={2026},
  doi={10.48550/arXiv.2510.05706}
}

@inproceedings{hu2023evolving,
  title={Evolving constrained reinforcement learning policy},
  author={Hu, C. and Pei, J. and Liu, J. and Yao, X.},
  booktitle={IJCNN},
  year={2023},
  doi={10.1109/IJCNN54540.2023.10191982}
}

@article{kushwaha2025survey,
  title={A survey of safe reinforcement learning and constrained mdps: A technical survey on single-agent and multi-agent safety},
  author={Kushwaha, A. and Ravish, K. and Lamba, P. and Kumar, P.},
  journal={arXiv:2505.17342},
  year={2025}, 
  doi={10.48550/arXiv.2505.17342}
}

@inproceedings{ma2022conservative,
  title={Conservative and adaptive penalty for model-based safe reinforcement learning},
  author={Ma, Y. and Shen, A. and Bastani, O. and Dinesh, J.},
  booktitle={AAAI},
  year={2022},
  doi={10.1609/aaai.v36i5.20478}
}
\bibliographystyle{iclr2027_conference}
\FloatBarrier

\newpage

\appendix

\FloatBarrier

\section{Setup and baseline details}
\label{apx:toy_setup}

\begin{table}[ht]
\centering
\caption{Base configuration. The planner and cascade parameters are those of Algorithm~\ref{alg:lex_cem}.}
\label{tab:appx_toy_config}
\begin{tabular}{lcl}
\toprule
\textbf{Parameter} & \textbf{Value} & \textbf{Description} \\
\midrule
$K$ (samples) & 1200 & CEM samples per iteration \\
$\ell$ (elites) & 24 & elites kept per iteration \\
$H$ (horizon) & 8 & planning horizon \\
iterations & 10 & CEM iterations \\
num\_act\_stepped & 4 & actions executed per replan \\
var\_scale & 1.0 & proposal variance scale \\
velocity cap & 1.2 & $\pm$ action clamp \\
goal tolerance & 0.5 & arrival radius \\
disk & $(1.5,1.5), r=1.4$ & obstacle \\
lane half-width & 0.8 & lane bounds \\
\bottomrule
\end{tabular}
\end{table}

\FloatBarrier
\section{Geometry robustness}
\label{apx:toy_ordering}

\begin{table}[ht]
\centering
\caption{Geometry robustness of the tier-order at the recommended configuration (Table~\ref{tab:appx_toy_config}) across 10 random seeds. The reference disk-on-lane geometry is re-run alongside three perturbations (an off-diagonal disk (center $(1.2,2.2)$, $r=1.3$), two overlapping disks, and a lane narrowed to half-width $0.5$) by generalizing the exact signed-distance model to an arbitrary list of disks. \emph{R} = seeds reaching the goal; \emph{V$_\uparrow$} = seeds where the \emph{higher}-priority geometry constraint went negative; \emph{V$_\downarrow$} = seeds where the lower-priority constraint went negative; \emph{Steps} = mean episode steps. For \emph{task\_last} the sacrificed tier is the task (both geometry constraints are protected, so \emph{V$_\downarrow$} is blank), and for \emph{plain} the cascade is off and neither constraint is protected.}
\label{tab:toy_geometry}
\begin{tabular}{ll|ccc|c}
\toprule
\textbf{Geometry} & \textbf{Config} & \textbf{R} & \textbf{V$_\uparrow$} & \textbf{V$_\downarrow$} & \textbf{Steps} \\
\midrule
base (reference) & \tt{task\_obstacle\_lane} & 10 & 0 & 10 & 46.4 \\
 & \tt{task\_lane\_obstacle} & 10 & 0 & 10 & 40.0 \\
 & \tt{task\_last} & 0 & 0 & --- & 160 \\
 & \tt{plain} & 9 & --- & --- & 74.8 \\
off-diagonal disk & \tt{task\_obstacle\_lane} & 10 & 0 & 0 & 42.0 \\
 & \tt{task\_lane\_obstacle} & 10 & 0 & 4 & 34.0 \\
 & \tt{task\_last} & 0 & 0 & --- & 160 \\
 & \tt{plain} & 10 & --- & --- & 54.8 \\
two overlapping disks & \tt{task\_obstacle\_lane} & 8 & 2 & 10 & 70.8 \\
 & \tt{task\_lane\_obstacle} & 10 & 0 & 10 & 38.4 \\
 & \tt{task\_last} & 0 & 0 & --- & 160 \\
 & \tt{plain} & 9 & --- & --- & 48.4 \\
tight lane ($0.5$) & \tt{task\_obstacle\_lane} & 2 & 0 & 10 & 153 \\
 & \tt{task\_lane\_obstacle} & 8 & 1 & 10 & 60.8 \\
 & \tt{task\_last} & 0 & 0 & --- & 160 \\
 & \tt{plain} & 10 & --- & --- & 73.2 \\
\bottomrule
\end{tabular}
\end{table}

\begin{figure}[!t]
    \centering
    \def\svgwidth{\linewidth}
    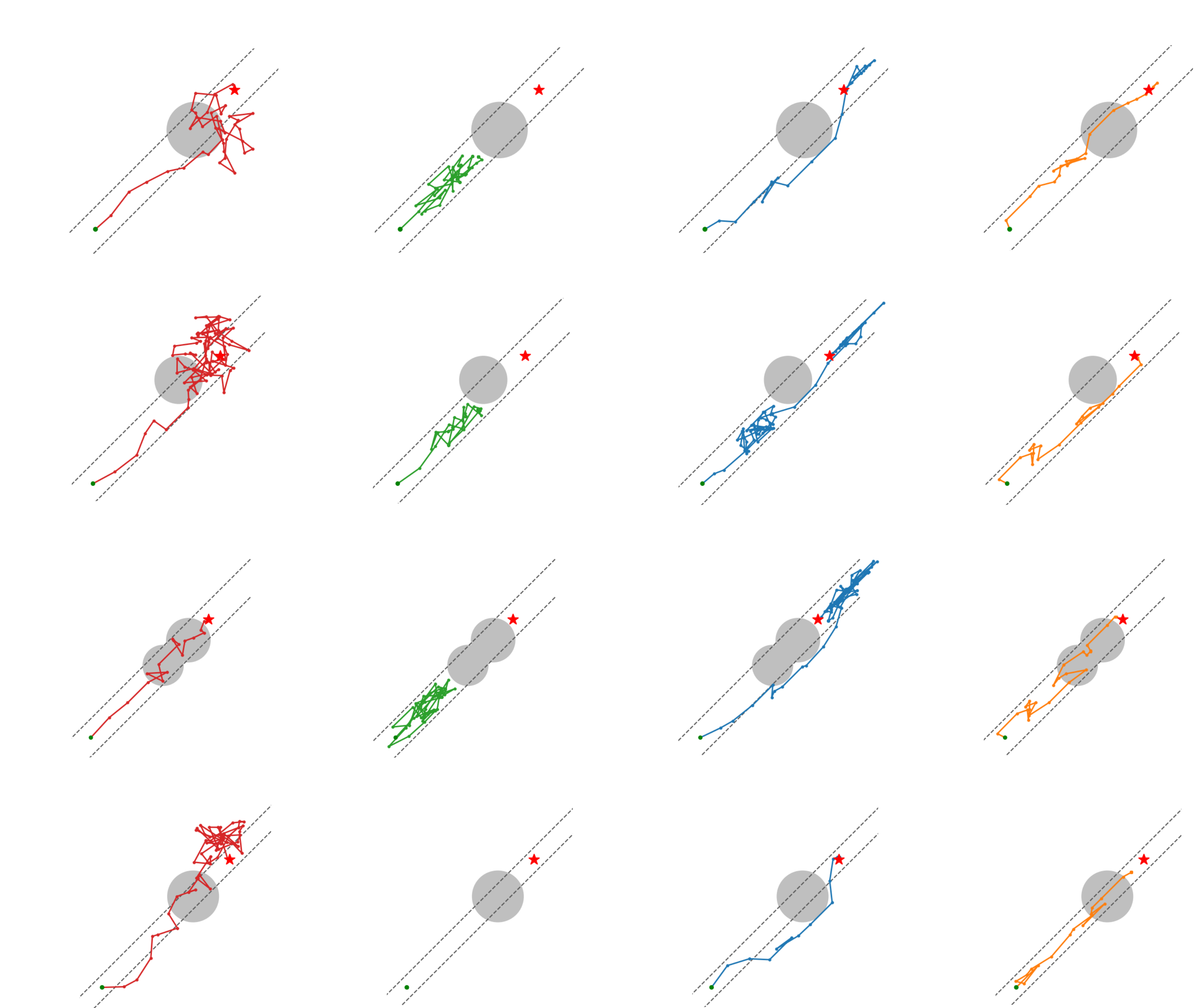
    \caption{Visualization of trajectories for different constraint configurations: each row is a geometry, each column a configuration. The shaded disks are the obstacles, the dashed lines bound the lane are the lane boundary, the green point is the start and the red star is the goal. Across every geometry \emph{plain} (first column) smashes through at least one constraint, \emph{task\_last} (second column) stalls preserving both, \emph{task\_obstacle\_lane} (third column) arcs around the obstacle sacrificing the lane while the obstacle stays clean, and \emph{task\_lane\_obstacle} (fourth column) drives straight sacrificing the obstacle while the lane stays clean.}
    \label{fig:toy_geometry}
\end{figure}

\FloatBarrier
\clearpage
\section{TierCEM hyperparameter search}
\label{apx:toy_sweep}

Table~\ref{tab:toy_sweep} varies one hyperparameter at a time over seeds 0--4. The best results found for TierCEM were at variance 1.00, 1200 samples, 24 elites, 8 horizon, 10 iterations and commit length 4, where both experiments reach the goal in all evaluated seeds with zero high-priority violations.

\begin{table}[h]
\centering
\small
\setlength{\tabcolsep}{2.5pt}
\caption{Grid Search of the mechanism analysis over 5 random seeds as one hyperparameter varies around the base configuration of Table~\ref{tab:appx_toy_config}. Each row lists the full hyperparameter tuple $(K, \ell, H, \mathrm{iters}, \mathrm{commit}, \sigma)$; bold marks the single value varied away from the base. \emph{R} = seeds reaching the goal; \emph{V$_\downarrow$} = seeds where the low-priority geometry constraint went negative (lane for \emph{task\_obstacle\_lane}, obstacle for \emph{task\_lane\_obstacle}); \emph{V$_\uparrow$} = seeds where the higher-priority constraint went negative.}
\label{tab:toy_sweep}
\begin{tabular}{ll|ccc|ccc}
\toprule
\textbf{Setting} & \textbf{Hyperparameters $(K,\ell,H,\mathrm{it},\mathrm{cm},\sigma)$} & \multicolumn{3}{c}{\tt{task\_obstacle\_lane}} & \multicolumn{3}{c}{\tt{task\_lane\_obstacle}} \\
 & & \textbf{R} & \textbf{V$_\downarrow$} & \textbf{V$_\uparrow$} & \textbf{R} & \textbf{V$_\downarrow$} & \textbf{V$_\uparrow$} \\
\midrule
base & $(1200,\,24,\,8,\,10,\,4,\,0.75)$ & 2 & 2 & 0 & 0 & 0 & 0 \\
samples 600 & $(\mathbf{600},\,24,\,8,\,10,\,4,\,0.75)$ & 2 & 2 & 0 & 1 & 1 & 0 \\
samples 2400 & $(\mathbf{2400},\,24,\,8,\,10,\,4,\,0.75)$ & 0 & 0 & 0 & 0 & 0 & 0 \\
samples 4800 & $(\mathbf{4800},\,24,\,8,\,10,\,4,\,0.75)$ & 0 & 0 & 0 & 0 & 0 & 0 \\
elites 12 & $(1200,\,\mathbf{12},\,8,\,10,\,4,\,0.75)$ & 0 & 0 & 0 & 1 & 1 & 0 \\
elites 48 & $(1200,\,\mathbf{48},\,8,\,10,\,4,\,0.75)$ & 0 & 0 & 0 & 0 & 0 & 0 \\
elites 96 & $(1200,\,\mathbf{96},\,8,\,10,\,4,\,0.75)$ & 0 & 0 & 0 & 0 & 0 & 0 \\
var scale 0.50 & $(1200,\,24,\,8,\,10,\,4,\,\mathbf{0.50})$ & 0 & 0 & 0 & 0 & 0 & 0 \\
var scale 1.00 & $(1200,\,24,\,8,\,10,\,4,\,\mathbf{1.00})$ & 5 & 5 & 0 & 5 & 5 & 0 \\
horizon 4 & $(1200,\,24,\,\mathbf{4},\,10,\,4,\,0.75)$ & 0 & 0 & 2 & 0 & 0 & 0 \\
horizon 6 & $(1200,\,24,\,\mathbf{6},\,10,\,4,\,0.75)$ & 0 & 0 & 0 & 0 & 0 & 0 \\
horizon 12 & $(1200,\,24,\,\mathbf{12},\,10,\,4,\,0.75)$ & 4 & 5 & 0 & 5 & 5 & 0 \\
iter 5 & $(1200,\,24,\,8,\,\mathbf{5},\,4,\,0.75)$ & 0 & 0 & 0 & 0 & 0 & 0 \\
iter 20 & $(1200,\,24,\,8,\,\mathbf{20},\,4,\,0.75)$ & 0 & 0 & 0 & 1 & 1 & 0 \\
stepped 2 & $(1200,\,24,\,8,\,10,\,\mathbf{2},\,0.75)$ & 0 & 0 & 0 & 0 & 0 & 0 \\
stepped 8 & $(1200,\,24,\,8,\,10,\,\mathbf{8},\,0.75)$ & 5 & 5 & 4 & 5 & 5 & 0 \\
\bottomrule
\end{tabular}
\end{table}

\clearpage
\FloatBarrier
\section{Baselines grid search}
\label{apx:toy_softfilter}

\begin{table}[h]
\centering
\small
\setlength{\tabcolsep}{4pt}
\caption{The Soft-Constraint CEM baseline, sweeping the scalar weight $\lambda$ (same hyperparameters as Table~\ref{tab:toy_softbaseline} across 20 random seeds). ``Reach" is the number of seeds reaching the goal, and V$_\mathrm{obs}$ and V$_\mathrm{lane}$ are the number of seeds where the forbidden disk and the lane are violated respectively. At every $\lambda \in [0.05, 2]$ the disk is clipped in all 20 seeds, while the lane is violated throughout; the only $\lambda$ that stops the disk from being clipped makes the planner stall (reach drops to 2/20 at $\lambda = 5$).}
\label{tab:toy_softbaseline_lambda}
\begin{tabular}{lccc}
\toprule
$\lambda$ & \textbf{Reach} & \textbf{V$_\mathrm{obs}$} & \textbf{V$_\mathrm{lane}$} \\
\midrule
0.05 & 20/20 & 20 & 5  \\
0.1  & 20/20 & 20 & 3  \\
0.2  & 20/20 & 20 & 7  \\
0.5  & 20/20 & 20 & 12 \\
1.0  & 20/20 & 20 & 14 \\
2.0  & 20/20 & 20 & 16 \\
5.0  & 2/20  & 3  & 3  \\
\bottomrule
\end{tabular}
\end{table}

\begin{table}[h]
\centering
\small
\setlength{\tabcolsep}{2.5pt}
\caption{The results of the hierarchical scalarized CEM (same hyperparameters as Table~\ref{tab:toy_softbaseline} across 20 random seeds). Each row is a pair $(\lambda_\mathrm{O}, \lambda_\mathrm{L})$, and \texttt{task\_obstacle\_lane} is run with that pair and \texttt{task\_lane\_obstacle} with the reversed pair, meaning the first weight is always associated with the highest priority constraint of each of the configurations. ``Reach" is the number of seeds reaching the goal, V$_\uparrow$ and V$_\downarrow$ are the number of seeds where the highest and lower priorities are violated respectively, and ``Steps" is the average number of steps executed to complete the episode.}
\label{tab:toy_softbaseline_lambda_perconstraint}
\begin{tabular}{lcccclcccc}
\toprule
 & \multicolumn{4}{c}{\tt{task\_obstacle\_lane}} & & \multicolumn{4}{c}{\tt{task\_lane\_obstacle}} \\
$(\lambda_\mathrm{O}, \lambda_\mathrm{L})$ (ratio) & \textbf{Reach} & \textbf{V$_\uparrow$} & \textbf{V$_\downarrow$} & \textbf{Steps} & & \textbf{Reach} & \textbf{V$_\uparrow$} & \textbf{V$_\downarrow$} & \textbf{Steps} \\
\midrule
(0.5, 0.02), 25:1 & 20/20 & 0 & 20 & 38 & & 20/20 & 0 & 20 & 16 \\
(1.0, 0.05), 20:1 & 20/20 & 1 & 20 & 41 & & 20/20 & 1 & 20 & 23 \\
(0.5, 0.05), 10:1 & 20/20 & 1 & 20 & 35 & & 20/20 & 0 & 20 & 25 \\
(1.0, 0.1), 10:1 & 20/20 & 1 & 20 & 22 & & 20/20 & 1 & 20 & 34 \\
(2.0, 0.2), 10:1 & 20/20 & 0 & 20 & 32 & & 20/20 & 2 & 20 & 35 \\
(5.0, 0.5), 10:1 & 20/20 & 2 & 20 & 40 & & 19/20 & 5 & 20 & 57 \\
(0.5, 0.1), 5:1 & 20/20 & 0 & 20 & 29 & & 20/20 & 6 & 20 & 34 \\
(0.5, 0.2), 2.5:1 & 20/20 & 1 & 20 & 30 & & 20/20 & 15 & 20 & 38 \\
(0.5, 0.35), 1.4:1 & 19/20 & 0 & 20 & 39 & & 20/20 & 20 & 17 & 35 \\
\bottomrule
\end{tabular}
\end{table}

\begin{figure}[h]
    \centering
    \def\svgwidth{0.7\textwidth}
    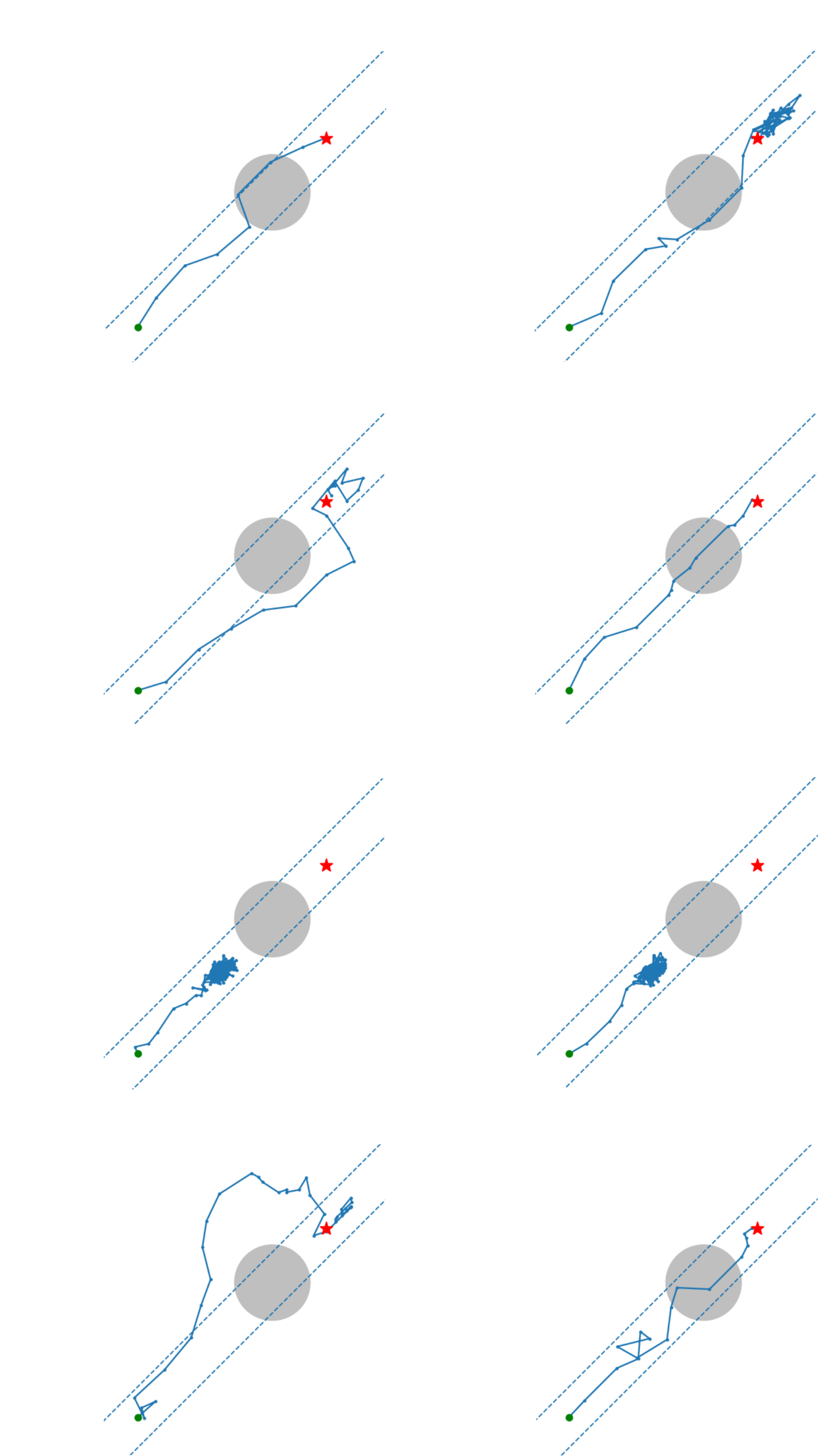
    \caption{The soft-constraint baseline versus the lexicographic cascade on the mechanism analysis environment (rows: Soft-constraint CEM at $\lambda = 0.5$, the hierarchical scalarized CEM at the tuned 25:1 ratio of  Table~\ref{tab:toy_softbaseline_lambda_perconstraint}, the Constrained Cross Entropy (CCE), and TierCEM; columns: the two tier orders). The Soft-Constraint CEM carries a single scalar weight and cuts the forbidden disk (negative \texttt{min\_obs}); TierCEM protects the higher-priority tier and sacrifices only the lower one. The hierarchical scalarized CEM, given per-constraint weights in the correct 25:1 ratio, also protects the higher-priority tier at this seed, but the ratio that achieves this must be tuned per environment whereas TierCEM encodes the ordering directly. CCE stalls to a safe state but never completes the task.}
    \label{fig:toy_softbaseline}
\end{figure}

\FloatBarrier

\section{TierCEM under inertial bodies}
\label{apx:toy_inertial}

A pure hierarchy does not guarantee a tier to become empty because a no-move action keeps the current state fixed in a safe state. This is a controllability assumption, as it only holds for a position-controlled; that is, kinematic, body whose action is a velocity, but not for a body with momentum, whose action is an acceleration. 

\paragraph{Setup.} This appendix repeats the environment of Section~\ref{sec:toy_progress}. The kinematic baseline uses a single integrator. The state is the position $x_t \in \mathbb{R}^2$ and the action $a_t$ is a velocity command clamped element-wise to the range $[-A_{\max},\, A_{\max}]$. The rollout is $  x_{t+1} = x_t + \min\,(A_{\max},\, \max(a_t,\, {-A_{\max}})) \Delta t$ (we use $\Delta t = 1$ for simplicity), with $A_{\max} = 1.2$.

The inertial body has state $(x, y, v_x, v_y)$ and action $a \in \mathbb{R}^2$. 
The rollout uses Euler integration with $\Delta t = 1$:
\begin{align}
    x_{t+1} &= x_t + v_t, \\
    v_{t+1} &= v_t + \mathrm{sat}(a_t), 
\end{align}
where $\mathrm{sat}(a_t)$ denotes element-wise saturation to $[-A_{\max},\, A_{\max}]$ and 
$A_{\max}=1.2$. For the "brake-only" case we further restrict $a_t \in [-A_{\max}, 0]^2$, 
so the body can decelerate to zero but cannot reverse its velocity. Priority order is defined as (i) obstacle and (ii) lane boundary; there is no task-progress tier. See Figure~\ref{fig:toy_inertial}.

.

\paragraph{Results.} Because the body cannot hold still under the zero action, a tier can empty  when no plan keeps the body inside the constraint over the horizon. The stopping distance is what determines which constraint is sacrificed. From speed $s$, braking at the maximum rate leaves a residual velocity $s - k\,A_\mathrm{max}\,dt$, which determines the distance advanced along the lane, meaning the stopping distance is  $E(s) = \frac{s^2}{2A_\mathrm{max}}$. When $E(s) \le$ gap, the body can stop before the disk and arc around it to avoid it, sacrificing the lane boundary. When $E(s) >$ gap, there is no time to stop and it goes through the obstacle, as the first priority tier produces an empty set of candidates (see Table~\ref{tab:toy_inertial}).

\begin{table}[h]
\centering
\small
\setlength{\tabcolsep}{3pt}
\caption{Zero-motion escape routes are only available to kinematic bodies. For inertial bodies we report stopping distance $E(s) = s^2 / (2A_\mathrm{max})$ with $A_\mathrm{max}=1.2$.
``Reach" is successes / 20 seeds. ``min obs" / ``min lane" are the minimum signed distances to the forbidden disk and lane boundary. Negative = constraint violation. Results are across 20 random seeds.}
\label{tab:toy_inertial}
\begin{tabular}{llcccc}
\toprule
\textbf{Body} & \textbf{Speed} $s$ & $E(s)$ & \textbf{Reach} & \textbf{min obs} & \textbf{min lane} \\
\midrule
\multicolumn{6}{l}{\textit{Kinematic}} \\
kinematic & 0.4 & --- & 0/20 & $+0.59$ & $+0.31$ \\
kinematic & 1.4 & --- & 0/20 & $+0.54$ & $+0.31$ \\
kinematic & 2.0 & --- & 0/20 & $+0.64$ & $+0.25$ \\
\midrule
\multicolumn{6}{l}{\textit{Inertial, gap = 1.60}} \\
inertial & 0.4 & 0.07 & 16/20 & $+0.14$ & $-9.23$ \\
inertial & 1.4 & 0.82 & 20/20 & $+0.14$ & $-8.37$ \\
inertial & 2.0 & 1.67 & 19/20 & $-0.16$ & $-7.24$ \\
inertial & 2.5 & 2.60 & 20/20 & $-0.24$ & $-7.60$ \\
inertial & 3.0 & 3.75 & 19/20 & $-1.32$ & $-4.51$ \\
\midrule
\multicolumn{6}{l}{\textit{Inertial, gap = 0.60}} \\
inertial & 0.4 & 0.07 & 18/20 & $+0.12$ & $-8.25$ \\
inertial & 1.4 & 0.82 & 19/20 & $-0.10$ & $-8.19$ \\
inertial & 1.8 & 1.35 & 18/20 & $-1.29$ & $-5.28$ \\
inertial & 2.0 & 1.67 & 20/20 & $-1.33$ & $-6.68$ \\
\bottomrule
\end{tabular}
\end{table}

\begin{figure}[h]
    \centering
    \def\svgwidth{\linewidth}
    \begingroup%
  \makeatletter%
  \providecommand\color[2][]{%
    \errmessage{(Inkscape) Color is used for the text in Inkscape, but the package 'color.sty' is not loaded}%
    \renewcommand\color[2][]{}%
  }%
  \providecommand\transparent[1]{%
    \errmessage{(Inkscape) Transparency is used (non-zero) for the text in Inkscape, but the package 'transparent.sty' is not loaded}%
    \renewcommand\transparent[1]{}%
  }%
  \providecommand\rotatebox[2]{#2}%
  \newcommand*\fsize{\dimexpr\f@size pt\relax}%
  \newcommand*\lineheight[1]{\fontsize{\fsize}{#1\fsize}\selectfont}%
  \ifx\svgwidth\undefined%
    \setlength{\unitlength}{544.08643828bp}%
    \ifx\svgscale\undefined%
      \relax%
    \else%
      \setlength{\unitlength}{\unitlength * \real{\svgscale}}%
    \fi%
  \else%
    \setlength{\unitlength}{\svgwidth}%
  \fi%
  \global\let\svgwidth\undefined%
  \global\let\svgscale\undefined%
  \makeatother%
  \begin{picture}(1,0.33803301)%
    \lineheight{1}%
    \setlength\tabcolsep{0pt}%
    \put(0,0){\includegraphics[width=\unitlength,page=1]{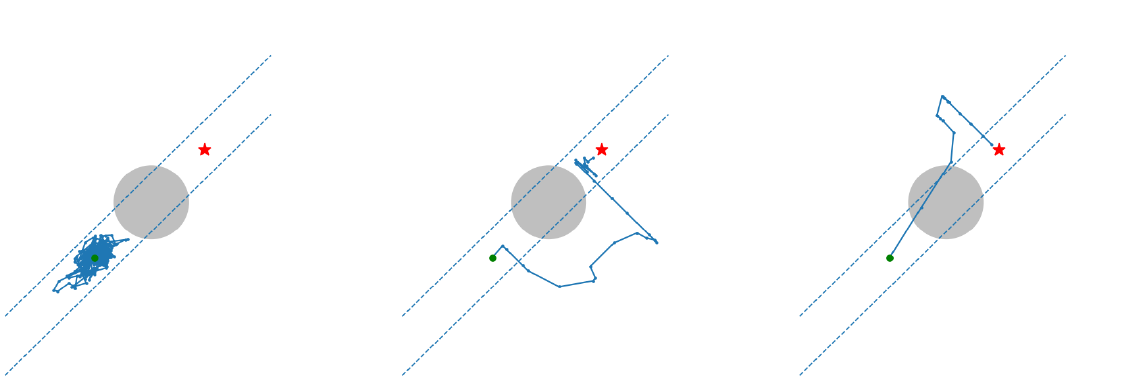}}%
    \put(0.14048359,0.31288999){\color[rgb]{0,0,0}\makebox(0,0)[t]{\lineheight{1.25}\smash{\begin{tabular}[t]{c}\textbf{kinematic}\end{tabular}}}}%
    \put(0.47107137,0.31289001){\color[rgb]{0,0,0}\makebox(0,0)[t]{\lineheight{1.25}\smash{\begin{tabular}[t]{c}\textbf{inertial $E(s) <$ gap}\end{tabular}}}}%
    \put(0.84010978,0.31269303){\color[rgb]{0,0,0}\makebox(0,0)[t]{\lineheight{1.25}\smash{\begin{tabular}[t]{c}\textbf{inertial $E(s) >$ gap}\end{tabular}}}}%
  \end{picture}%
\endgroup%

    \caption{The kinematic control never reaches the goal (left), meaning no tier ever empties and, therefore, the planner stalls and both geometry constraints stay positive. The inertial body  sacrifices the lane constraint in order to complete the goal (center), unless the gap between the starting point and the obstacle is too narrow for the initial speed $s$ (right).}
    \label{fig:toy_inertial}
\end{figure}

\FloatBarrier

\section{Imperfect constraint signals}
\label{apx:imperfect}

\begin{figure}[h]
    \centering
    \def\svgwidth{\linewidth}
    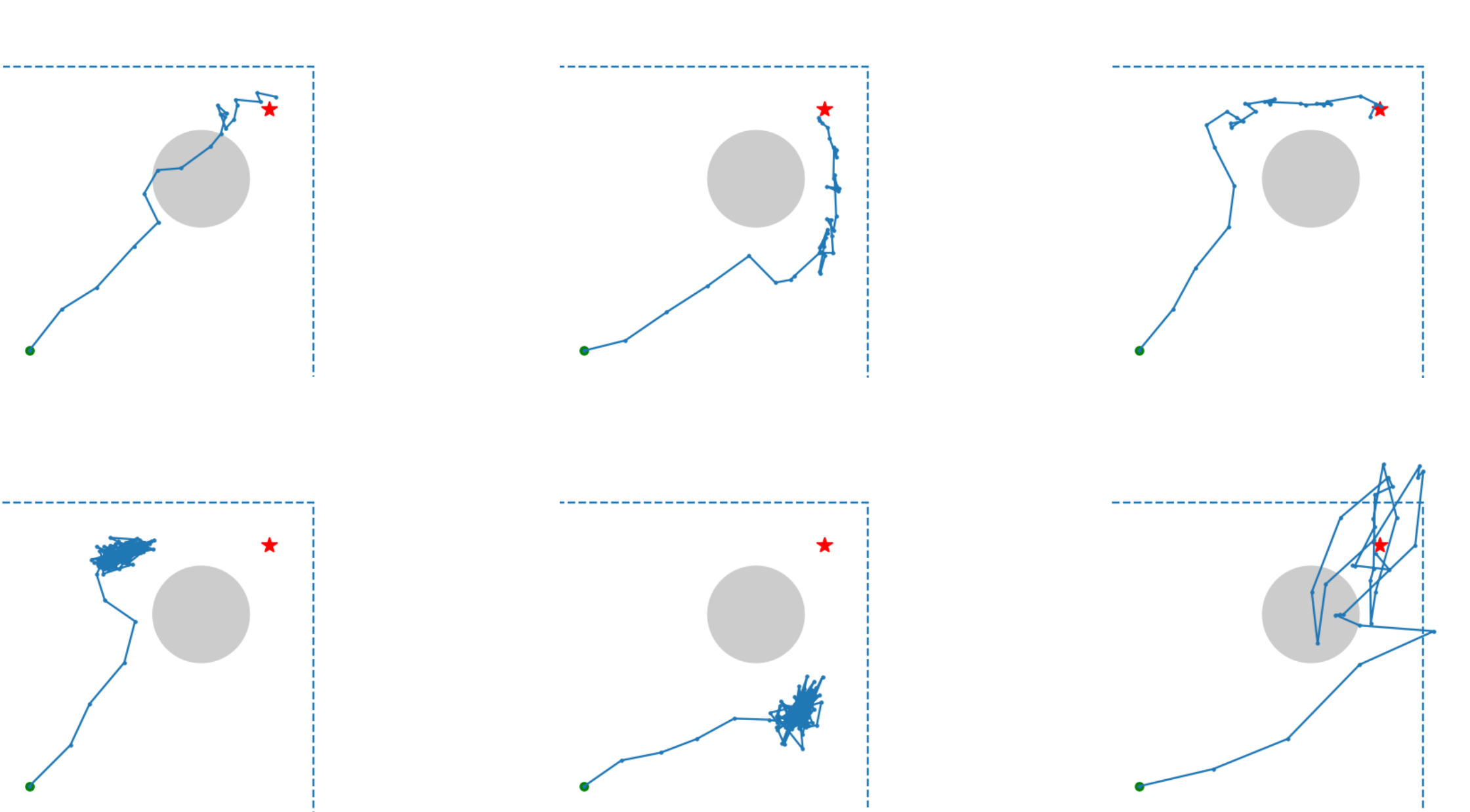
    \caption{Imperfect constraint signals. \emph{plain} cuts through the disk; \emph{oracle} and \emph{mlp\_head} reach the goal safely; \emph{mlp\_shifted} and \emph{noisy\_head} stall short, hallucinating phantom unsafe regions; \emph{model\_mismatch} plans a safe arc but the executed path clips the disk and goes outside the workspace boundary occasionally.}
    \label{fig:toy_first}
\end{figure}

This section isolates the effect of the constraint signal's quality. We use a second environment using the same setup as in Section~\ref{sec:toy} but replacing the lane boundary by a squared workspace boundary. We define six different scenarios: plain (conventional CEM), oracle (exact, noise-free margin functions), MLP heads (an MLP predicting the margin function for each constraint), MLP shifted (the same MLP trained only on the half of the workspace left of the obstacle, so the goal region and part of the forbidden disk must be extrapolated), noisy head (exact distances corrupted by $\mathcal{N}(0, 0.35)$ noise), and model mismatch (exact margins, but the optimizer's internal dynamics advances time with an 80\% speed error). Figure~\ref{fig:toy_first} and Table~\ref{tab:toy_heads} show that errors in the margins produce over-conservative behavior, while errors in the dynamics model produce dangerous false-confidence, violating multiple times each constraint.

\begin{table}[h]
\centering
\caption{Second controlled environment (with workspace-boundary tier). \emph{min dist} is the minimum distance to the goal along the executed path, \emph{min obs} / \emph{min ws} the minimum signed distance to the obstacle / workspace boundary (negative = violated). Results are across 5 random seeds.}
\label{tab:toy_heads}
\begin{tabular}{lcccc}
\toprule
\textbf{Scenario} & \textbf{min dist} & \textbf{min obs} & \textbf{min ws} & \textbf{Verdict} \\
\midrule
\texttt{plain}         & 0.21 & $-0.85$ & $+0.64$ & reaches goal, crashes through \\
\texttt{oracle}        & 0.23 & $+0.67$ & $+0.85$ & reaches goal, always safe \\
\texttt{mlp\_head}     & 0.23 & $+0.68$ & $+0.80$ & reaches goal, always safe \\
\texttt{mlp\_shifted}  & 3.31 & $+0.61$ & $+0.71$ & never arrives, over-cautious \\
\texttt{noisy\_head}   & 3.94 & $+1.00$ & $+1.25$ & never arrives, over-cautious \\
\texttt{model\_mismatch} & 0.23 & $-0.27$ & $-0.45$ & arrives, violates (3/5 seeds) \\
\bottomrule
\end{tabular}
\end{table}

\FloatBarrier
 
\section{MLP margin predictor details and hyperparameters}
\label{apx:mlp_hyper}

\paragraph{Targets and temporal alignment.} The deployed model is a shared-backbone binary classifier with one output for each competing constraint; that is, the forbidden disk and the lane boundary. The exact signed margins are positive in safe states:

\begin{align}
    m_{\mathrm{obs}}(s_t) =\lVert p_{\mathrm{ee},t}-c_{\mathrm{obs}}\rVert_2-r_{\mathrm{obs}}\,, \\
    m_{\mathrm{lane}}(s_t) =w_{\mathrm{lane}}-\lvert p^y_{\mathrm{ee},t}-y_{\mathrm{lane}}\rvert\,,
\end{align}

where $p_{\mathrm{ee},t}$ is the end-effector position, $c_{\mathrm{obs}}$ and $r_{\mathrm{obs}}$ are the forbidden-disk center and radius, and $w_{\mathrm{lane}}$ is the lane half-width. The binary target for constraint $n$ is:

\begin{equation}
        y_{t,n}=\mathbf{1}\!\left[\min_{j}m_n(s_{t+j})<0\right] , 
\end{equation}

which means that a predicted world-model transition is labeled unsafe if the end-effector violates the constraint at any of the environment states throughout the action horizon. Training uses six auto-regressive DINO-WM steps and the action rollouts are normalized to match the distribution the planner sees during inference.

\paragraph{Input and architecture.} Each input is a 384-dimensional mean-pooled latent predicted by the frozen DINO-WM. Input features are standardized per dimension using only the training split. The deployed shared classifier has dimensions $384 \rightarrow 256 \rightarrow 64 \rightarrow 2$, and each hidden linear layer is followed by LayerNorm, GELU, and dropout with probability $0.2$. The final linear layer produces two heads, ordered as obstacle and lane boundary, and a sigmoid converts each logit to $P_n(\mathrm{violation}\mid z_t)$. In total, this classifier contains 115,778 trainable parameters, while the DINO-WM remains frozen.

\paragraph{Optimization and model selection.} We optimize the classifier with unweighted binary cross-entropy; no class weights or resampling are applied. AdamW starts at a learning rate of $3\times10^{-4}$ with weight decay $10^{-4}$ over 200 epochs. We save and use the weights with the lowest validation binary cross-entropy and stop after 25 epochs without an improvement greater than $10^{-6}$.

\begin{table}[h]
\centering
\caption{Hyperparameters of the shared DINO-WM classifier used for the forbidden-zone and lane-boundary heads.}
\label{tab:safety_head_hyperparameters}
\small
\begin{tabular}{p{0.43\linewidth}p{0.49\linewidth}}
\toprule
\textbf{Hyperparameter} & \textbf{Value} \\
\midrule
Input dimension & 384 \\
Hidden dimensions & $(256,64)$ \\
Output dimension & 2 \\
Hidden activation & GELU \\
Normalization & LayerNorm after each hidden linear layer \\
Dropout & 0.2 \\
Trainable parameters & 115,778 \\
Maximum epochs & 200 \\
Batch size & 1024 \\
Optimizer & AdamW \\
Initial/minimum learning rate & $3\times10^{-4}$ / $10^{-6}$ \\
Learning-rate schedule & Cosine annealing \\
Weight decay & $10^{-4}$ \\
Training loss & Unweighted masked binary cross-entropy \\
Early-stopping patience & 25 epochs \\
Validation improvement tolerance & $10^{-6}$ \\
Train/validation/test split & $70\%/15\%/15\%$, by episode \\
Split seed / training seed & 42 / 42 \\
Input standardization & Yes, training statistics only \\
World-model horizon & 6 model steps \\
Violation decision threshold & $P_n(\mathrm{violation})\geq0.5$ \\
\bottomrule
\end{tabular}
\end{table}

\begin{figure}[h]
    \centering

    \begin{subfigure}[b]{\linewidth}
        \centering
        \def\svgwidth{0.8\linewidth}
        \begingroup%
  \makeatletter%
  \providecommand\color[2][]{%
    \errmessage{(Inkscape) Color is used for the text in Inkscape, but the package 'color.sty' is not loaded}%
    \renewcommand\color[2][]{}%
  }%
  \providecommand\transparent[1]{%
    \errmessage{(Inkscape) Transparency is used (non-zero) for the text in Inkscape, but the package 'transparent.sty' is not loaded}%
    \renewcommand\transparent[1]{}%
  }%
  \providecommand\rotatebox[2]{#2}%
  \newcommand*\fsize{\dimexpr\f@size pt\relax}%
  \newcommand*\lineheight[1]{\fontsize{\fsize}{#1\fsize}\selectfont}%
  \ifx\svgwidth\undefined%
    \setlength{\unitlength}{588.98535541bp}%
    \ifx\svgscale\undefined%
      \relax%
    \else%
      \setlength{\unitlength}{\unitlength * \real{\svgscale}}%
    \fi%
  \else%
    \setlength{\unitlength}{\svgwidth}%
  \fi%
  \global\let\svgwidth\undefined%
  \global\let\svgscale\undefined%
  \makeatother%
  \begin{picture}(1,0.5040416)%
    \lineheight{1}%
    \setlength\tabcolsep{0pt}%
    \put(0,0){\includegraphics[width=\unitlength,page=1]{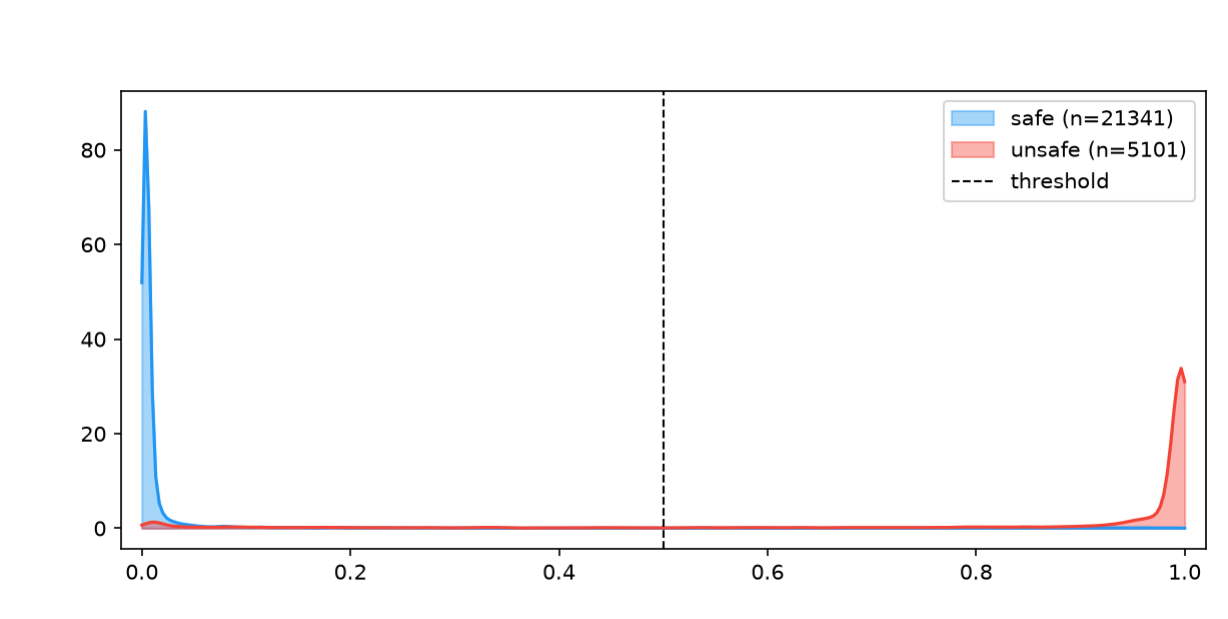}}%
    \put(0.52540428,0.46792284){\color[rgb]{0.30196078,0.30196078,0.30196078}\makebox(0,0)[t]{\lineheight{1.25}\smash{\begin{tabular}[t]{c}\textbf{Obstacle heads}\end{tabular}}}}%
    \put(0.54665776,0.00614802){\color[rgb]{0.30196078,0.30196078,0.30196078}\makebox(0,0)[t]{\lineheight{1.25}\smash{\begin{tabular}[t]{c}P(violation)\end{tabular}}}}%
    \put(0.02237167,0.23143771){\color[rgb]{0.30196078,0.30196078,0.30196078}\rotatebox{90.00691456}{\makebox(0,0)[t]{\lineheight{1.25}\smash{\begin{tabular}[t]{c}Density\end{tabular}}}}}%
  \end{picture}%
\endgroup%

        \label{fig:fz_pusht}
    \end{subfigure}
    \hfill
    \vspace{1em}
    \begin{subfigure}[b]{0.8\linewidth}
        \centering
        \def\svgwidth{\linewidth}
        \begingroup%
  \makeatletter%
  \providecommand\color[2][]{%
    \errmessage{(Inkscape) Color is used for the text in Inkscape, but the package 'color.sty' is not loaded}%
    \renewcommand\color[2][]{}%
  }%
  \providecommand\transparent[1]{%
    \errmessage{(Inkscape) Transparency is used (non-zero) for the text in Inkscape, but the package 'transparent.sty' is not loaded}%
    \renewcommand\transparent[1]{}%
  }%
  \providecommand\rotatebox[2]{#2}%
  \newcommand*\fsize{\dimexpr\f@size pt\relax}%
  \newcommand*\lineheight[1]{\fontsize{\fsize}{#1\fsize}\selectfont}%
  \ifx\svgwidth\undefined%
    \setlength{\unitlength}{590.13939565bp}%
    \ifx\svgscale\undefined%
      \relax%
    \else%
      \setlength{\unitlength}{\unitlength * \real{\svgscale}}%
    \fi%
  \else%
    \setlength{\unitlength}{\svgwidth}%
  \fi%
  \global\let\svgwidth\undefined%
  \global\let\svgscale\undefined%
  \makeatother%
  \begin{picture}(1,0.50387164)%
    \lineheight{1}%
    \setlength\tabcolsep{0pt}%
    \put(0,0){\includegraphics[width=\unitlength,page=1]{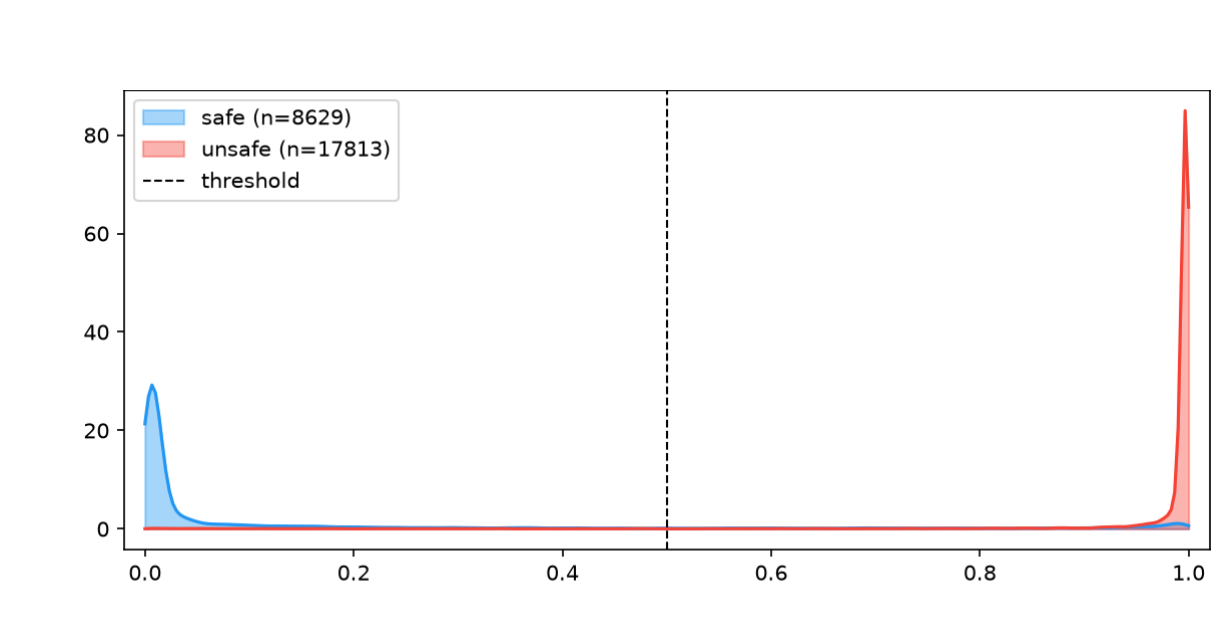}}%
    \put(0.54346114,0.48069783){\color[rgb]{0.30196078,0.30196078,0.30196078}\makebox(0,0)[t]{\lineheight{1.25}\smash{\begin{tabular}[t]{c}\textbf{Lane heads}\end{tabular}}}}%
    \put(0.53727314,0.00613604){\color[rgb]{0.30196078,0.30196078,0.30196078}\makebox(0,0)[t]{\lineheight{1.25}\smash{\begin{tabular}[t]{c}P(violation)\end{tabular}}}}%
    \put(0.02298946,0.2509533){\color[rgb]{0.30196078,0.30196078,0.30196078}\rotatebox{91.53863241}{\makebox(0,0)[t]{\lineheight{1.25}\smash{\begin{tabular}[t]{c}Density\end{tabular}}}}}%
  \end{picture}%
\endgroup%

        \label{fig:wb_pusht}
    \end{subfigure}
    \begin{subfigure}[b]{0.7\linewidth}
        \centering
        \def\svgwidth{\linewidth}
        \begingroup%
  \makeatletter%
  \providecommand\color[2][]{%
    \errmessage{(Inkscape) Color is used for the text in Inkscape, but the package 'color.sty' is not loaded}%
    \renewcommand\color[2][]{}%
  }%
  \providecommand\transparent[1]{%
    \errmessage{(Inkscape) Transparency is used (non-zero) for the text in Inkscape, but the package 'transparent.sty' is not loaded}%
    \renewcommand\transparent[1]{}%
  }%
  \providecommand\rotatebox[2]{#2}%
  \newcommand*\fsize{\dimexpr\f@size pt\relax}%
  \newcommand*\lineheight[1]{\fontsize{\fsize}{#1\fsize}\selectfont}%
  \ifx\svgwidth\undefined%
    \setlength{\unitlength}{504.00000865bp}%
    \ifx\svgscale\undefined%
      \relax%
    \else%
      \setlength{\unitlength}{\unitlength * \real{\svgscale}}%
    \fi%
  \else%
    \setlength{\unitlength}{\svgwidth}%
  \fi%
  \global\let\svgwidth\undefined%
  \global\let\svgscale\undefined%
  \makeatother%
  \begin{picture}(1,0.85714289)%
    \lineheight{1}%
    \setlength\tabcolsep{0pt}%
    \put(0,0){\includegraphics[width=\unitlength,page=1]{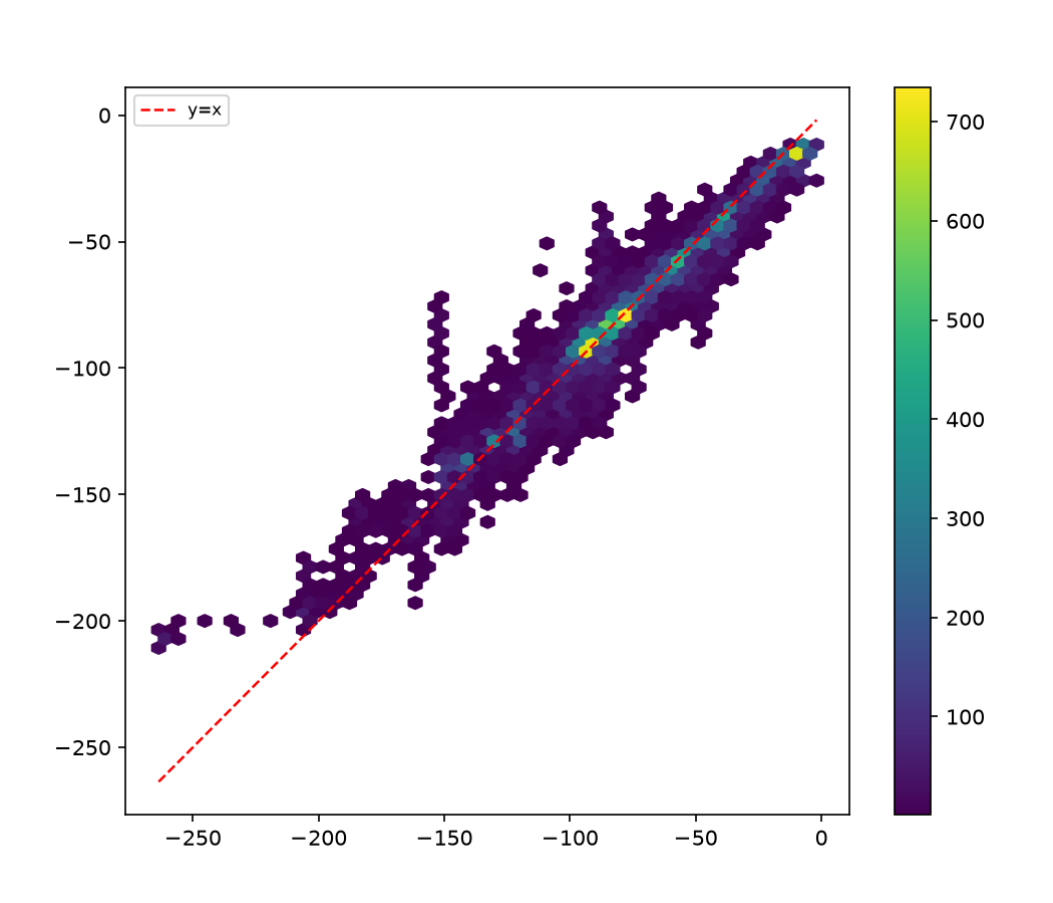}}%
    \put(0.49909084,0.80060694){\color[rgb]{0,0,0}\makebox(0,0)[t]{\lineheight{1.25}\smash{\begin{tabular}[t]{c}\textbf{Predicted vs True block to goal distance}\end{tabular}}}}%
    \put(0.46148509,0.02041104){\color[rgb]{0,0,0}\makebox(0,0)[t]{\lineheight{1.25}\smash{\begin{tabular}[t]{c}True (px)\end{tabular}}}}%
    \put(0.03283967,0.42877083){\color[rgb]{0,0,0}\rotatebox{90.103535}{\makebox(0,0)[t]{\lineheight{1.25}\smash{\begin{tabular}[t]{c}Predicted (px)\end{tabular}}}}}%
  \end{picture}%
\endgroup%

    \label{fig:task_head}
    \end{subfigure}
    \caption{First two rows: resulting KDE of the MLP prediction for the forbidden disk and the lane boundary; third row: the predicted block to goal distance versus the true block to goal distance.}
    \label{fig:pusht_mlp}
\end{figure}

\FloatBarrier

\section{Task-progress MLP hyperparameters}
\label{apx:task_progress_mlp}

\paragraph{Input and architecture.} The input is the 384-dimensional mean-pooled visual latent predicted by the frozen DINO-WM. The regressor's hidden layer dimensions are $384 \rightarrow 32 \rightarrow 32 \rightarrow 1$, with a linear layer, LayerNorm, GELU activation, and dropout with probability of $0.5$ after each hidden layer. The output layer is linear and has no output activation. This model contains 13,537 trainable parameters; DINO-WM remains frozen.

\paragraph{Optimization and model selection.} We minimize mean squared error in standardized target space with AdamW. Table~\ref{tab:task_progress_hyperparameters} shows the complete configuration. We keep the checkpoint with the lowest validation MSE and stop when validation loss has not improved by at least $10^{-6}$ for 20 consecutive epochs. 

\begin{table}[h]
\centering
\caption{Hyperparameters of the DINO-WM task-progress MLP. Values are the defaults used by the fixed-conflict training pipeline.}
\label{tab:task_progress_hyperparameters}
\small
\begin{tabular}{p{0.43\linewidth}p{0.49\linewidth}}
\toprule
\textbf{Hyperparameter} & \textbf{Value} \\
\midrule
Input dimension & 384 \\
Hidden dimensions & $(32,32)$ \\
Output dimension & 1 \\
Hidden activation & GELU \\
Normalization & LayerNorm after each hidden linear layer \\
Dropout & 0.5 \\
Maximum epochs & 200 \\
Batch size & 1024 \\
Optimizer & AdamW \\
Learning rate & $10^{-3}$ \\
Weight decay & $10^{-2}$ \\
Training loss & Mean squared error \\
Early-stopping patience & 20 epochs \\
Validation improvement tolerance & $10^{-6}$ \\
Train/validation/test split & $70\%/15\%/15\%$, by episode \\
Split seed / training seed & 0 / 0 \\
Input and target standardization & Yes, training statistics only \\
World-model horizon & 6 model steps \\
\bottomrule
\end{tabular}
\end{table}

\newpage
\section{Behavior when constraints are jointly satisfiable}
\label{apx:noconflict}

This appendix reports a Push-T variant that complements the deliberately fixed-conflict experiment of Sections~\ref{sec:pusht_proprio} and~\ref{sec:pusht}. Here, the constraints are an obstacle and a workspace boundary, which admit useful trajectories satisfying both constraints, and the goal is sampled from the expert dataset. The main experiment of Sections~\ref{sec:pusht_proprio} and~\ref{sec:pusht} instead uses a fixed goal and an obstacle that spans a horizontal lane, forcing an ordered relaxation for task completion. 

In this new variant, we do not expect reversing the tiers to change behavior, as both can be satisfied simultaneously while being successful in the task. Its purpose is to establish the scope of the mechanism. We aim to prove that sequential filtering is a useful mechanism also when constraints do not conflict with each other. We train an MLP with a shared backbone with two heads, one for each constraint, as done in the main experiments, and we do not use a task progress margin. 

\begin{figure}[t]
\centering
\begin{subfigure}[b]{0.19\textwidth}
    \centering
    \includegraphics[width=\textwidth]{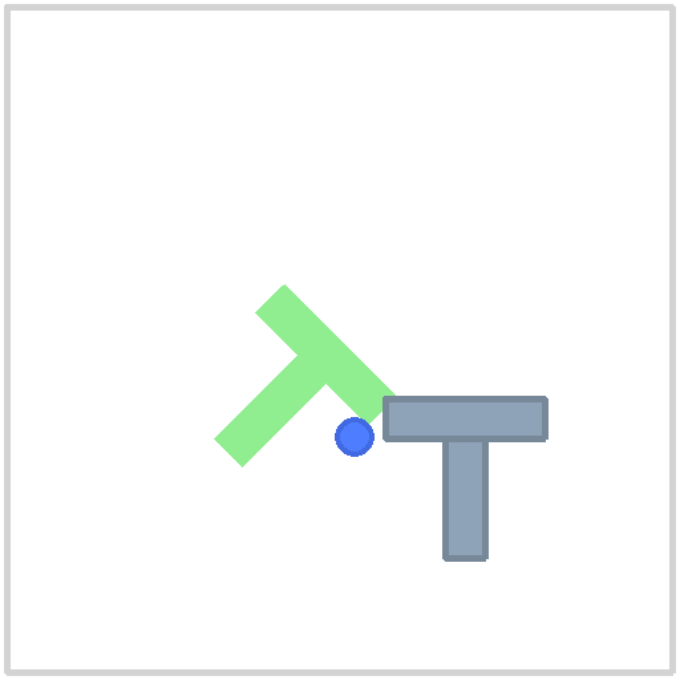}
    \caption{}
    \label{fig:pusht_env_sub1}
\end{subfigure}
\hfill
\begin{subfigure}[b]{0.19\textwidth}
    \centering
    \includegraphics[width=\textwidth]{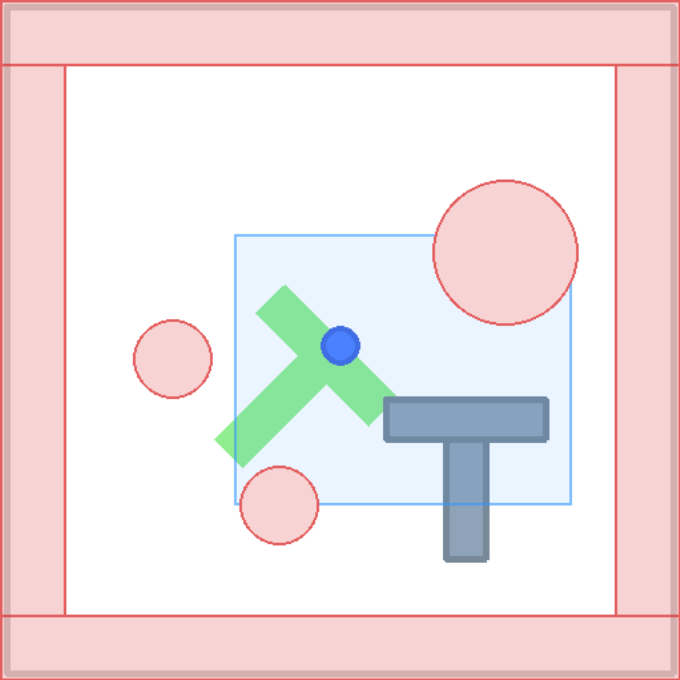}
    \caption{}
    \label{fig:pusht_env_sub2}
\end{subfigure}
\hfill
\begin{subfigure}[b]{0.19\textwidth}
    \centering
    \includegraphics[width=\textwidth]{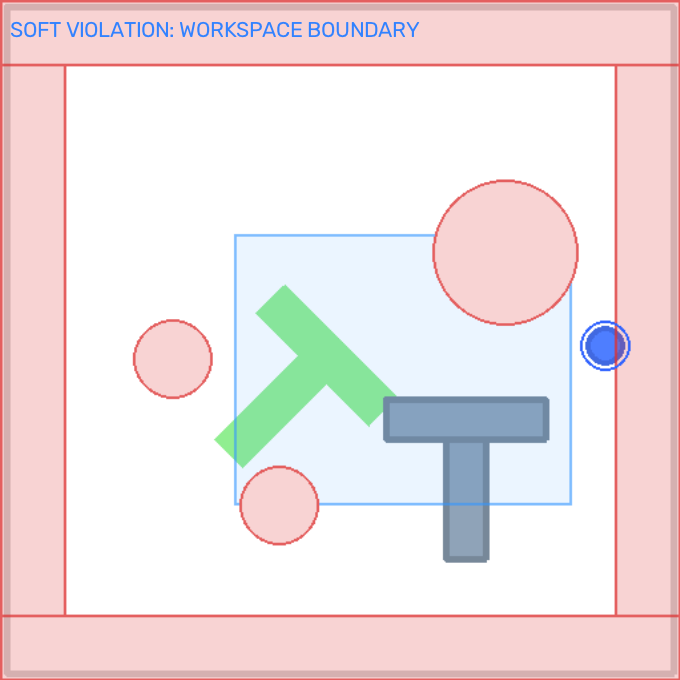}
    \caption{}
    \label{fig:pusht_env_sub4}
\end{subfigure}
\hfill
\begin{subfigure}[b]{0.19\textwidth}
    \centering
    \includegraphics[width=\textwidth]{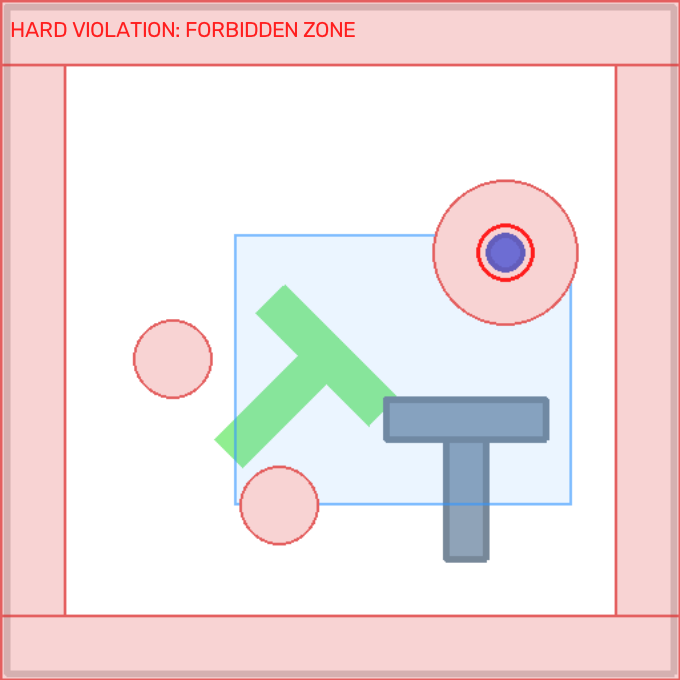}
    \caption{}
    \label{fig:pusht_env_sub5}
\end{subfigure}
\caption{Image representation samples of the Push-T environment.~\ref{fig:pusht_env_sub1} shows the image observation used as an input for the world models. For the rest of images, red zones are forbidden regions, while the blue zone represents the workspace boundary. When a violation happens, it is marked around the end-effector with a circle surrounding it (red for an obstacle violation, and blue for a workspace boundary violation). ~\ref{fig:pusht_env_sub2} shows an example where all constraints are satisfied,~\ref{fig:pusht_env_sub4} shows a workspace boundary violation, and~\ref{fig:pusht_env_sub5} shows an obstacle violation.}
\label{fig:pusht_env}
\end{figure}

We compare scalarized planners (Tables~\ref{tab:soft-mean} and~\ref{tab:hier-scalarized}) and TierCEM (Table~\ref{tab:lex-cem}) under the expert-dataset goal. All results average 100 episodes over seeds 1--5. These sweeps are used to show the non-conflict regime rather than as additional evidence for constraint relaxation. In this setup, we use as $R_{\mathrm{task}}^{(k)}$ the negative MSE of the terminal latent and the goal latent to match the trajectory-selection criterion used by DINO-WM~\citep{pmlr-v267-zhou25t} and keep the comparison focused on the effect of the priority ordering mechanism rather than on a different task ranking.

\begin{table}[h]
\centering
\caption{Soft-constraint CEM with scalarized penalty, evaluated for 100 episodes across 5 random seeds.}
\label{tab:soft-mean}
\begin{tabular}{lccccc}
\toprule
$\lambda$ & O (\%) & WB (\%) & Pos Dist (px) & Angle Error (\textdegree) \\
\midrule
0.05 & $10.9 \pm 1.7$ & $35.0 \pm 3.4$ & $15.2 \pm 3.9$ & $12.6 \pm 3.9$  \\
0.1  & $10.4 \pm 1.4$ & $31.8 \pm 3.5$ & $15.9 \pm 2.3$ & $12.8 \pm 2.4$  \\
0.3  & $10.2 \pm 1.8$ & $27.7 \pm 2.6$ & $16.7 \pm 2.8$ & $14.3 \pm 2.9$ \\
0.5  & $10.3 \pm 2.3$ & $26.6 \pm 3.2$ & $17.3 \pm 3.1$ & $13.6 \pm 3.1$  \\
1.0  & $11.0 \pm 2.6$ & $23.1 \pm 2.4$ & $19.2 \pm 3.2$ & $15.6 \pm 4.1$  \\
\bottomrule
\end{tabular}
\end{table}

\begin{table}[h]
\centering
\caption{Hierarchical scalarized CEM, evaluated for 100 episodes across 5 random seeds.}
\label{tab:hier-scalarized}
\begin{tabular}{lcccc}
\toprule
$\lambda_{\text{O}}$ / $\lambda_{\text{WB}}$ & O (\%) & WB (\%) & Pos Dist (px) & Angle Error (\textdegree) \\
\midrule
0.25 / 0.01 & $11.2 \pm 1.7$ & $37.5 \pm 2.2$ & $15.6 \pm 2.4$ & $13.1 \pm 4.0$ \\
0.5  / 0.02 & $9.9  \pm 1.6$ & $34.8 \pm 2.2$ & $16.3 \pm 3.1$ & $13.7 \pm 4.2$ \\
1    / 0.04 & $10.1 \pm 1.6$ & $32.0 \pm 2.5$ & $16.7 \pm 3.3$ & $14.0 \pm 3.7$ \\
0.5  / 0.1  & $9.8  \pm 1.8$ & $29.2 \pm 2.9$ & $16.9 \pm 2.8$ & $13.7 \pm 3.3$ \\
\bottomrule
\end{tabular}
\end{table}

\FloatBarrier

\subsection{Number of TierCEM iterations sweep}
\label{apx:warmup_grid}

TierCEM can be activated after a number $X$ of unconstrained CEM iterations. Table~\ref{tab:lex-cem} sweeps the number of cascade iterations on the Push-T environment (100 episodes, seeds 1--5), with the remaining $30 - X$ iterations running as unconstrained CEM. Workspace-boundary violations fall monotonically from 41.0\% to 20.9\%, at the expense of task accuracy. It is possible therefore to choose the operating point on this trade-off while the priority ordering itself remains unchanged.

\begin{figure}[t]
    \centering
    \def\svgwidth{\linewidth}
    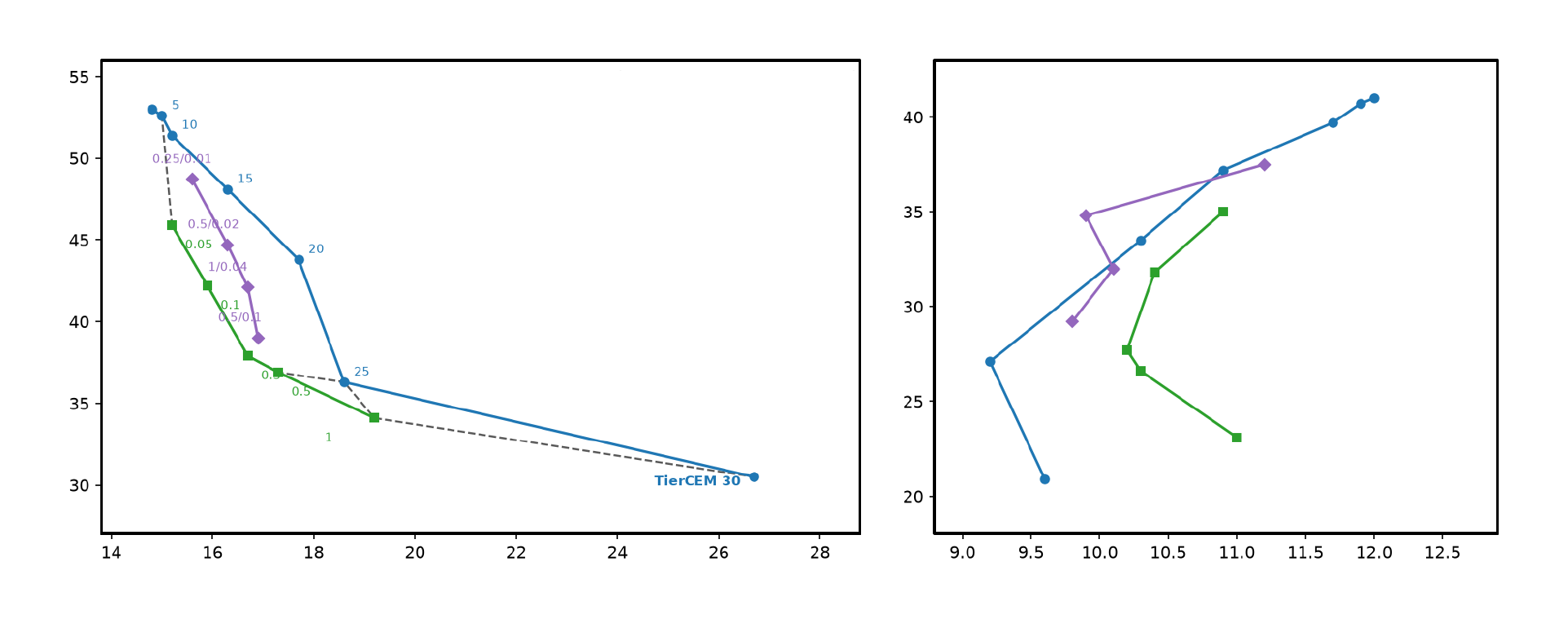
    \caption{Pareto presentation of the Push-T sweeps of Tables~\ref{tab:soft-mean}, \ref{tab:hier-scalarized}, and the obstacle as highest priority column of~\ref{tab:lex-cem} (means over 100 episodes across 5 random seeds). These results do not use OOD detection. (a) Total violation rate against position error, the dashed step line is the Pareto front. (b) The same sweeps in the ordering plane.}
    \label{fig:pushT_pareto}
\end{figure}

Figure~\ref{fig:pushT_pareto} plots the sweeps of Tables \ref{tab:soft-mean}, \ref{tab:hier-scalarized} and~\ref{tab:lex-cem} together. The ordering plane of Figure~\ref{fig:pushT_pareto} shows where TierCEM is beneficial with respect to the other baselines. TierCEM at 30 iterations reaches $9.6\%$ forbidden-zone and $20.9\%$ workspace-boundary violations, below every scalarized point on both axes, without a weight expressing the trade-off, at the $26.7$ px position error that is visualized in the aggregate trade-off.

Table~\ref{tab:lex-cem} additionally compares the priority order used in Section~\ref{sec:pusht} (obstacle above workspace boundary) against the reversed order (workspace boundary above obstacle). The two columns are nearly identical across all cascade iteration counts. This is because, unlike the deliberately conflicting geometry of Section~\ref{sec:toy}, here the constraints are not in conflict. TierCEM's role on the task is therefore not to choose which constraint to relax, but to enforce both constraints.

\begin{table}[t]
\centering
\caption{Grid search over the number of lexicographic cascade iterations on Push-T without OOD detection, evaluated for 100 episodes across 5 random seeds, comparing the normal priority order against reversed priorities. Each row runs $X = 30 - \mathrm{iterations}$ unconstrained CEM warm-up iterations followed by the remaining cascade iterations.}
\label{tab:lex-cem}
\begin{adjustbox}{max width=\columnwidth}
\begin{tabular}{lcccc|cccc}
\toprule
& \multicolumn{4}{c}{\textbf{Obstacle as highest priority}} & \multicolumn{4}{c}{\textbf{Workspace boundary as highest priority}} \\
\cmidrule(lr){2-5} \cmidrule(lr){6-9}
Cascade iterations & O (\%) & WB (\%) & Pos Dist (px) & Angle Error (\textdegree) & O (\%) & WB (\%) & Pos Dist (px) & Angle Error (\textdegree) \\
\midrule
0  & $12.0 \pm 1.1$ & $41.0 \pm 2.0$ & $14.8 \pm 3.2$ & $12.3 \pm 3.6$ & $12.0 \pm 1.1$ & $41.0 \pm 2.0$ & $14.8 \pm 3.2$ & $12.3 \pm 3.6$ \\
5  & $11.9 \pm 1.2$ & $40.7 \pm 1.9$ & $15.0 \pm 3.1$ & $12.5 \pm 3.6$ & $11.9 \pm 1.4$ & $40.7 \pm 2.2$ & $14.9 \pm 3.6$ & $12.5 \pm 3.9$ \\
10 & $11.7 \pm 1.1$ & $39.7 \pm 2.2$ & $15.2 \pm 2.6$ & $12.6 \pm 3.4$ & $11.7 \pm 1.3$ & $39.7 \pm 2.6$ & $14.9 \pm 3.2$ & $12.4 \pm 3.9$ \\
15 & $10.9 \pm 1.0$ & $37.2 \pm 1.7$ & $16.3 \pm 2.5$ & $13.1 \pm 3.9$ & $11.2 \pm 1.4$ & $37.6 \pm 2.6$ & $16.1 \pm 3.2$ & $12.8 \pm 4.7$ \\
20 & $10.3 \pm 1.7$ & $33.5 \pm 2.9$ & $17.7 \pm 2.9$ & $13.5 \pm 3.1$ & $10.3 \pm 1.9$ & $33.0 \pm 2.5$ & $18.2 \pm 2.6$ & $14.0 \pm 3.4$ \\
25 & $9.2  \pm 1.6$ & $27.1 \pm 2.4$ & $18.6 \pm 3.2$ & $14.8 \pm 2.6$ & $9.2  \pm 1.7$ & $26.9 \pm 2.9$ & $18.9 \pm 3.5$ & $15.0 \pm 3.1$ \\
30 & $9.6  \pm 1.3$ & $20.9 \pm 2.8$ & $26.7 \pm 4.1$ & $23.0 \pm 4.9$ & $9.3\pm 1.9$ & $19.7 \pm 3.2$ & $28.0 \pm 4.0$ & $24.4 \pm 4.4$ \\
\bottomrule
\end{tabular}
\end{adjustbox}
\end{table}

\FloatBarrier

\subsection{Optional third tier}
\label{apx:ood}

We additionally implemented an exploratory three-tier ordering, OOD $\succ$ obstacle $\succ$ workspace boundary. The OOD detector was implemented as a diffusion model trained on safe DINO-WM latents that scores candidates with the SCOPED statistic~\citep{barkley2025scoped}. We include this only as evidence that the cascade accepts a third, heterogeneous predicate without changing its selection rule. OOD detection is not a contribution of this paper and is disabled in all quantitative results above. Results are shown in Table~\ref{tab:combined}.

\begin{table}[ht]
\centering
\caption{Effect of OOD detection on Push-T performance across methods, DINO-WM latents, averaged across 5 random seeds. ``O'' and ``WB'' denote obstacle and workspace boundary constraint violation rates. ``Pos'' and ``Angle'' denote average position distance and orientation error of the T-block relative to the goal.}
\label{tab:combined}
\small
\setlength{\tabcolsep}{2.5pt}
\renewcommand{\arraystretch}{1.1}
\begin{adjustbox}{max width=\columnwidth}
\begin{tabular}{@{}llcccc|cccc@{}}
\toprule
& & \multicolumn{4}{c}{\textbf{No OOD}} &
    \multicolumn{4}{c}{\textbf{With OOD}} \\
\cmidrule(lr){3-6} \cmidrule(lr){7-10}
\textbf{Method} & \textbf{Config} &
\textbf{O} (\%) & \textbf{WB} (\%) &
\textbf{Pos.} (px) & \textbf{Angle} ($^\circ$) &
\textbf{O} (\%) & \textbf{WB} (\%) &
\textbf{Pos.} (px) & \textbf{Angle} ($^\circ$) \\
\midrule
\textbf{Baseline CEM} & --
& $12.0\!\pm\!1.1$ & $41.0\!\pm\!2.0$ & $14.8\!\pm\!3.2$ & $12.3\!\pm\!3.6$
& $8.9\!\pm\!1.6$ & $32.6\!\pm\!4.6$ & $16.0\!\pm\!2.4$ & $13.4\!\pm\!2.5$ \\
\textbf{TierCEM} & 30 iters
& $9.6\!\pm\!1.3$ & $20.9\!\pm\!2.8$ & $26.7\!\pm\!4.1$ & $23.0\!\pm\!4.9$
& $7.6\!\pm\!1.0$ & $19.3\!\pm\!2.4$ & $24.7\!\pm\!3.0$ & $21.4\!\pm\!3.2$ \\
\textbf{Hard-constraint flat tier} & 30 iters
& $9.9\!\pm\!2.5$ & $21.0\!\pm\!3.9$ & $26.6\!\pm\!4.2$ & $22.7\!\pm\!5.2$
& $7.4\!\pm\!1.5$ & $19.4\!\pm\!3.5$ & $24.8\!\pm\!2.6$ & $21.3\!\pm\!3.3$ \\
\textbf{CCE~\citep{wen2018constrained}} & --
& $8.7\!\pm\!1.6$ & $20.4\!\pm\!4.1$ & $26.1\!\pm\!1.5$ & $23.7\!\pm\!4.5$
& $8.4\!\pm\!1.2$ & $20.5\!\pm\!3.4$ & $23.8\!\pm\!2.8$ & $21.7\!\pm\!5.2$ \\
\textbf{Soft-constraint CEM} & $\lambda=0.5$
& $10.3\!\pm\!2.3$ & $26.6\!\pm\!3.2$ & $17.3\!\pm\!3.1$ & $13.6\!\pm\!3.1$
& $9.2\!\pm\!1.2$ & $23.2\!\pm\!3.2$ & $18.2\!\pm\!1.8$ & $14.2\!\pm\!3.2$ \\
\textbf{Hierarchical scalarized CEM}
& \makecell[t]{
$\lambda_{\mathrm{O}}=0.5$ \\
$\lambda_{\mathrm{WB}}=0.1$
}
& $9.8\!\pm\!1.8$ & $29.2\!\pm\!2.9$ & $16.9\!\pm\!2.8$ & $13.7\!\pm\!3.3$
& $9.5\!\pm\!1.7$ & $26.7\!\pm\!4.4$ & $17.5\!\pm\!2.0$ & $12.9\!\pm\!3.2$ \\
\bottomrule
\end{tabular}
\end{adjustbox}
\end{table}

\end{document}